\pdfoutput=1
\documentclass{article}
\usepackage[T1]{fontenc}
\usepackage{arxiv}

\usepackage{microtype}
\usepackage{hyperref}
\usepackage{url}
\usepackage{amsmath}
\usepackage{amsfonts}
\usepackage{amssymb}
\usepackage{booktabs}
\usepackage{graphicx}
\usepackage{subcaption}
\usepackage{float}
\graphicspath{{./}{./Figures/}}
\usepackage{algorithm}
\usepackage[noend]{algorithmic}
\usepackage{wrapfig}
\usepackage[normalem]{ulem}
\usepackage{pifont}
\usepackage{xspace}
\usepackage[table]{xcolor}
\usepackage{bm}
\usepackage{tcolorbox, listings}
\tcbuselibrary{skins, breakable, listings, raster}
\usepackage{multirow}
\usepackage{longtable}
\usepackage{array}
\usepackage{adjustbox}
\usepackage{accents}   %
\usepackage{booktabs}
\usepackage{tabularx}

\usepackage[dvipsnames]{xcolor}
\usepackage{titletoc}

\newcommand{\blfootnote}[1]{%
  \begingroup
  \renewcommand\thefootnote{}\footnote{#1}%
  \addtocounter{footnote}{-1}%
  \endgroup
}

\definecolor{bfayellow}{rgb}{1,1,0.6}
\definecolor{bfagreen}{rgb}{0.6,0.95,0.6}

\title{Physics of Agents: Statistical Mechanics Predicts Collective Behavior of AI Agents}
\author{%
  \begin{tabular}{cccc}
    Batu El$^{1\dag}$ & Jinhee Paeng$^{1}$ & Fatih Dinc$^{2}$ & Shiye Su$^{1}$ \\[0.6em] Mete Erdogan$^{1}$ &
    Aneesh Pappu$^{1}$  
    & Haotian Ye$^{1}$
     & Wanjia Zhao$^{1}$ 
  \end{tabular} \\[0.6em]
  \begin{tabular}{ccc}
   & Surya Ganguli$^{1}$ & James Zou$^{1\dag}$
  \end{tabular}
}
\affiliation{$^{1}$Stanford University \quad $^{2}$UC Santa Barbara \quad 
}

\begin{document}

\blfootnote{%
\centering
\begin{tabular}{llll}
Code & \href{https://github.com/batu-el/physics-of-agents}{github.com/batu-el/physics-of-agents} &
Online Appendix & \href{https://batu-el.github.io/physics-of-agents}{batu-el.github.io/physics-of-agents} \\
Dataset & \href{https://huggingface.co/physics-of-agents}{huggingface.co/physics-of-agents}
&
\multicolumn{2}{l}{{\dag} Correspondence to \texttt{\{batuel,jamesz\}@stanford.edu}}
\end{tabular}%
}

\maketitle

\vspace{-0.5cm}
\begin{abstract}
AI agents increasingly operate as part of interacting systems rather than in isolation. As agents exchange information and jointly make decisions, their interactions can improve collective reasoning but may also produce herding, polarization, or amplify shared biases.
Understanding and predicting these collective dynamics is therefore important for designing effective and aligned multi-agent systems. Here, we study over 10,000 communities of language-model agents that repeatedly exchange messages and revise their opinions across objective mathematics questions and subjective political statements. Despite substantial diversity in possible behavior, the individual and group dynamics can be represented by three characteristic regimes: indifference, polarization, and consensus. AI agents start indifferent and build conviction as they interact. On objective questions, communication improves collective accuracy, while on subjective questions it often drifts group opinions toward the right in the political spectrum. We explain these observations with a statistical-mechanics formalism in which agents stochastically favor lower social pressure. Given only initial opinions, our model predicts individual trajectories, outperforms all standard baselines, generalizes to unseen community graphs, and reproduces the observed group archetype distributions. Our fitted model parameters reveal the mechanics underlying our key observations: i) communities operate below the critical social temperature, which explains conviction buildup; ii) attractive ties outweigh repulsive ones, which favors consensus; and iii) agents holding the correct answer exert the strongest pull, which drives truth-seeking. Overall, our results demonstrate that collective behavior of AI agents, like that of other complex systems, follows compact and predictive dynamical laws.

\end{abstract}

\begin{figure}[h]
    \centering
\includegraphics[width=0.99\textwidth]{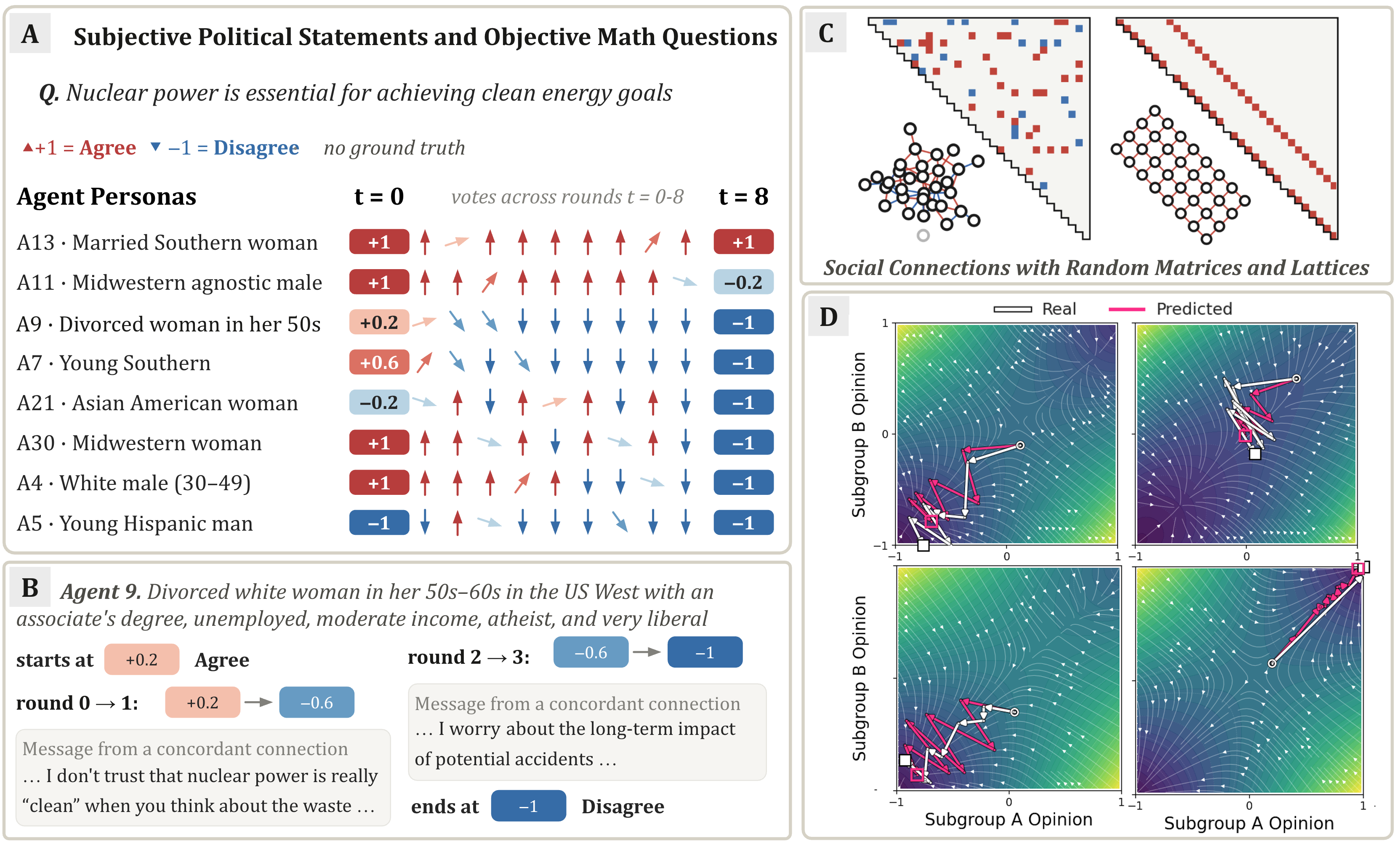}
\caption{\textbf{A. Opinion Updates.} Opinions of eight agents over eight rounds of message exchange, from their initial opinion at $t=0$ to their final opinion at $t=8$; agents start out mostly agreeing on one side and end up on the other. \textbf{B. Message Exchange.} Messages received by the agents and the opinion changes that followed. \textbf{C. Communication Networks.} Two of the four families of communication networks (random matrices, square lattices) shown as the signed adjacency $J$ as a heatmap in one corner triangle (blue $-1$, white $0$, red $+1$) and as a node-link diagram. See Appendix \ref{apdx:communication_networks} \textbf{D. Prediction.}  Each panel is one
group of $N$ language-model agents interacting about a question over eight rounds. We split the agents into two subgroups based on their opinion at $t=0$ (see Appendix \ref{apdx:f1} for details). White arrows are the real (observed) opinion changes, and the pink lines are discrete opinion updates predicted with our fitted model rolled out from the initial opinions at $t=0$. The background shows the energy in colors and predicted flow, which describe continuous limit, in streamlines.}
     \label{fig:1_intro}
\end{figure}

\section{Introduction}\label{sec:intro}
AI agents increasingly operate as part of interacting systems rather than in isolation. In scientific research, software engineering, and other complex domains, multiple agents exchange findings and opinions, critique proposed solutions, and jointly refine decisions. Systems such as the Virtual Lab \citep{Swanson2024.11.11.623004} and EinsteinArena \citep{bianchi2026harnessingcollectiveintelligenceai} demonstrate that flexible teams of language-model agents can carry out substantial parts of the scientific discovery process, from literature synthesis and hypothesis generation to computational analysis and experimental design. Related multi-agent interactions are increasingly used for coding, planning, debate, and automated research \citep{du2023improvingfactualityreasoninglanguage,liang-etal-2024-encouraging,chen2023}.

Interacting agents may also become part of everyday life. Personal assistant agents could communicate on behalf of their users to schedule meetings, negotiate purchases, coordinate travel, allocate shared resources, or resolve competing preferences. In such settings, the behavior that matters is not only that of any one assistant but also the collective outcome produced by many assistants interacting with one another. These interactions can aggregate complementary information, but they can also produce herding, polarization, persistent disagreement, oscillation, or the amplification of shared biases. As the number and autonomy of deployed agents grow, understanding these collective dynamics becomes important for both the design of effective multi-agent systems and AI alignment. Given the agents, their initial states, and the network through which they communicate, can we predict how the system will evolve? When will interaction improve collective decisions, and when will it instead entrench errors or amplify undesirable biases?

To answer these questions, we study communities of language-model agents that hold opinions about a shared question, communicate over a social network, and repeatedly revise their views (Figure \ref{fig:1_intro}). Each agent is assigned a distinct persona or area of expertise, and pairs of agents are connected by either concordant or discordant relationships. We consider both objective mathematical questions with verifiable answers and subjective political statements without unique ground-truth answers. Across multiple language models, questions, communication networks, and episodes, we simulate around $~10,000$ agent communities and record how their individual and collective opinions evolve over repeated rounds of interaction (Section~\ref{sec:setup}).

Our empirical analysis reveals diverse but structured collective dynamics (Section~\ref{sec:observations}). At the level of individual agents, opinion trajectories fall into recurring archetypes, in which some agents remain frozen, some switch once, some reverse and later return to their original position, and others oscillate repeatedly (Section~\ref{sec:macrostates_microstates}). At the population level, communities exhibit three characteristic regimes: indifference, in which agents hold weak opinions; polarization, in which strongly committed agents divide into opposing camps; and consensus, in which agents converge toward the same position. Although communities commonly begin with weakly held opinions, interaction consistently increases conviction and progressively moves them toward more ordered states (Section~\ref{sec:obs_phase_char}). Analysis of communication patterns reveal that this ordering is not a simple strategy where the population locks in on the initial majority, instead communities frequently weaken or overturn their initial collective position. On objective questions, these changes improve collective accuracy on average, with initially incorrect majorities switching to the correct answer more often than initially correct majorities becoming incorrect. On subjective questions, however, communication can produce systematic ideological drift, demonstrating that interaction may also amplify directional biases inherited from the underlying language models (Section~\ref{sec:truth_seeking}).

To predict and interpret these behaviors, we introduce a theoretical model of interacting agents in which individuals update their opinions to minimize the social pressure defined over a social graph (Section~\ref{sec:method}). We represent each agent's opinion as a binary variable whose evolution is governed by two forces: i) an intrinsic bias term (referred to as ``intrinsic field'') that captures the agent's predisposition toward the question, and ii) an interaction term quantifying social pressure by capturing the influence exerted by neighboring agents. This construction yields a theoretical model, whose mathematics can be mapped to the well-known Ising model from statistical mechanics literature \citep{ising1925,glauber1963}. We use this machinery to derive a stochastic update rule for opinions in which the probability of an agent adopting either opinion depends on its persona, the question, and the opinions of the agents connected to it. We extend the Ising model with multiple interaction parameters, which we fit with intrinsic fields directly from observed opinion transitions.

Despite its simplicity, the resulting model accurately predicts the collective behavior of agents on held-out questions, substantially outperforms standard baselines, and generalizes from random communication graphs to previously unseen network families (Section~\ref{sec:results_prediction}). When rolled out from observed initial conditions, it also approximately reproduces the population-level distribution of collective outcomes (Section~\ref{sec:results_distributions}). The fitted dynamics further provide a quantitative interpretation of the empirical observations. The communities operate in an ordered, low-temperature regime, consistent with the observed buildup of conviction (Section~\ref{sec:results_temperature}). Concordant interactions are substantially stronger than discordant interactions, favoring consensus over persistent polarization. On objective questions, agents currently holding the correct answer exert greater fitted influence than agents holding an incorrect answer, providing a statistical account of the observed improvement in collective accuracy (Section~\ref{sec:results_truthseeking}).

Together, these results suggest that the collective behavior of interacting AI agents can be described by compact dynamical laws analogous to those used for other complex systems. Statistical mechanics provides a framework for identifying recurring regimes, forecasting population dynamics, and relating system-level outcomes to interpretable interaction parameters.

\section{Setup}\label{sec:setup}
We consider a population of $N$ language model (LM) agents that hold opinions about a shared question, communicate with each other, and revise their opinions through multiple rounds of interactions (Figure \ref{fig:1_intro}).

\textit{Personas and Questions.} The agents answer one of two kinds of questions. \emph{Objective questions} are competition mathematics problems from the MATH dataset \citep{hendrycks2021measuringmathematicalproblemsolving}, recast as binary multiple choice questions. For example, \textit{``Find the distance between the foci of the hyperbola
$x^2 - 6x - 4y^2 - 8y = 27$. A: $4\sqrt{5}$\quad B: $4\sqrt{10}$''}. Here an agent's persona encodes \emph{expertise}: each agent is shown the worked ground-truth solution to a different MATH problem, so agents differ in the expertise they bring to the shared question. \emph{Subjective questions} are political statements adapted from the Political Questions Dataset for LLM Bias Evaluation \citep{promptfoo_political_bias_2025}, which each agent either agrees ($+1$) or disagrees ($-1$) with. For example,
\textit{``Mandatory vaccination violates bodily autonomy''}. Here personas are short free-text profiles adapted from TWIN-2K-500 \citep{toubia2025twin2k500datasetbuildingdigital}, describing an individual's demographics, political alignment, and personality traits, so agents differ in preference. Appendix~\ref{apdx:tasks} gives the full construction of both datasets.

\textit{Agents.} Each agent $i$ carries a fixed persona $p_i$, which distinguishes it from other agents and introduces heterogeneity into the population. The persona reflects variation in personal preferences (subjective) and competence (objective).\footnote{For the agents answering subjective questions, personas are based on profiles of real individuals from TWIN-2K-500
\citep{toubia2025twin2k500datasetbuildingdigital}. For objective questions, we encode
$p_i$ as domain expertise using the ground truth answers from \citet{hendrycks2021measuringmathematicalproblemsolving}.} At each time step $t$, agent $i$ processes an input $x_i(t)$, which consists of messages the agent $i$ received from its connections, and samples an opinion and a message in natural language that expresses the agent's opinion.\footnote{We use $N=32$ agents with unique personas in our experiments unless stated otherwise.} 

\textit{Communication Network.} A group of $N$ agents is connected by a communication network that determines which agents are allowed to send and receive messages from one another and the nature of the interaction between any two connected agents. We encode this network as $J \in \{-1,0,+1\}^{N\times N}$, where
a social tie $J_{ij}=+1$ means that agent $i$ is \textit{friendly (concordant)} with agent $j$, $J_{ij}=-1$ means that agent $i$ is \textit{unfriendly (discordant)} with agent $j$, and $J_{ij}=0$ means that the two agents do not communicate. These social ties model settings where an agent is more likely to trust opinions or information from certain agents and less likely for other neighbors (e.g. they might be instructed to do so by the agent owners). This formulation also includes a special case where all the edges have the same sign, so that there is no asymmetry in how agents regard each other. 
\footnote{We use symmetric $J$ with $0$ diagonal in all our experiments.} We generate the graphs $J$ from four families, which are demonstrated in Figure \ref{fig:2_Setup_Js} and described in Appendix~\ref{apdx:communication_networks}.

\paragraph{Procedure.}
All agents advance in synchronous rounds $t = 0,1,\dots,T$:

\textit{Sampling Opinions and Messages.} At each round $t$, agent $i$ expresses an answer to the question as a binary vote  $o_i(t) \sim \pi_i(\;\cdot\; | x_i(t))$ where $o_i(t) \in \{+1, -1\}$, $\pi$ is the language model, $\pi_i$ denotes conditioning on agent $i$'s persona, and $x_i(t)$ represents the messages and the question that are in the context of agent $i$ at timestep $t$. To reduce the sampling noise, we sample each agent's vote $K=5$ times and obtain $o_{i,k}(t)$ for $k = 1,\dots,K$. Then, we set $\bar{o}_i(t) = \frac{1}{K}\sum_{k=1}^K o_{i,k}(t)$ to be the opinion, so it takes one of six evenly spaced values in $[-1, 1]$. During message composition, the agent chooses a state following $s_i(t) = \text{sign}(\bar{o}_i(t))$ and samples a short message in natural language, $\omega_i(t) \sim \pi_i(\;\cdot\; | x_i(t), s_i(t) )$. The message contains its current position and provides one supporting reason.

\textit{Inbox Routing.} Each message is filed in the receiver's inbox according to the sign of their connection. If $J_{ij}=+1$ ($J_{ij}=-1$) and agent $i$ sends a message to agent $j$, the message arrives on a $+1$ ($-1$) edge and lands in the receiver's \textit{messages from friendly connections} (\textit{messages from unfriendly connections}) inbox (see Figure \ref{fig:opinion_prompt_examples} and Appendix \ref{apdx:prompts})

\textit{Opinion update.} Each agent considers its persona, the question, and its two inboxes, and re-generates its opinion to compose its message for the next turn. The updates are Markovian, \textit{i.e.},  the inboxes are refreshed at the start of each round and contain only the latest messages. Overall, influence propagates each round through the messages and opinions exchanged along the graph defined by
$J$. At $t=0$, each agent forms an opinion from its persona ($\{p_i\}_{i=1}^n$) and the question ($q$) alone.

Given the personas, the question, and the communication network, the system evolves into a terminal state ($s_1(T)$,  $s_2(T)$, \dots, $s_N(T)$), where $t=T$ is the final timestep.\footnote{We set $T=8$ in our experiments, unless stated otherwise.}
$$
(\{p_i\}_{i=1}^N,\ q,\ J)\ \longmapsto\ (s_1(t),  s_2(t), \dots, s_N(t)) \quad \text{for} \quad t = 0,1,\dots,T.
$$
We examine the behavior of this system when answering both objective and subjective questions. These two settings differ both in the questions posed and in the way personas are constructed as we described above. We provide further details on the questions, the personas, and the dataset construction for both settings in Appendix~\ref{apdx:tasks}. The backbone of the agents is a language model. We list the models we use and the embeddings we compute in Appendix~\ref{apdx:models}. For simplicity, all the communications and updates are synchronized in each round in the simulation\footnote{Notably, synchronized updates can lead to qualitative disagreements between discretized and continuous updates, \textit{e.g.}, emergence of oscillations in an energy-based system, as shown in Figure \ref{fig:1_intro}.}. We also extend the setup to asynchronous updates with similar results, which we investigate in Appendix \ref{apdx:continuous-extension}. Refer to Table \ref{tab:glossary} for a full glossary.

While our setup is simple, it captures the structure of multi-agent systems used in practice. For example, Mixture-of-Agents \citep{wang2024mixtureofagentsenhanceslargelanguage} has each agent read the outputs of all agents from the previous round, which is analogous to our procedure with a fully connected, all-friendly $J$ and depth $T$. Similarly, debate and ensemble methods correspond to other choices of $J$ and $T$ \citep{Swanson2024.11.11.623004,  du2023improvingfactualityreasoninglanguage}, and self-consistency-style methods correspond to $J$ being the identity, with every agent seeing only its own response from the previous turn \citep{wang2023selfconsistencyimproveschainthought}. Many agentic systems also use critique mechanisms, where an agent is asked to point out weaknesses of another agent's response, sometimes with agents told to disagree on purpose \citep{liang-etal-2024-encouraging}, similar to $J_{ij} = -1$. Meanwhile, systems used for scientific discovery, such as AlphaEvolve \citep{novikov2025alphaevolvecodingagentscientific}, ShinkaEvolve \citep{lange2025shinkaevolveopenendedsampleefficientprogram}, and SimpleTES \citep{ye2026structuredscalingaidiscovery}, fit a relaxation of the same template with candidate solutions as messages and the sampling policy as a dynamic $J$.

\section{Empirical Characterization of Agents' Collective Dynamics}\label{sec:observations}

In this section, we analyze $9,600$ simulated communities\footnote{$4$ models, $(40+20)$ objective and subjective questions, $10$ random graphs and lattices, and $4$ independent episodes for each group trajectory in each configuration. We use the word episodes to refer to the independent random samples with the same configuration. Together with the out-of-distribution generalization and asynchronous-update experiments, the full dataset comprises over $10{,}000$ groups, which is the figure quoted in the abstract.} to empirically characterize the individual- and group-level dynamics of the agents. Throughout this section, we conduct our analysis using the continuous values of opinions, $\bar{o}_i(t) \in [ -1, +1]$.

\subsection{Behavior of Groups and Individuals }\label{sec:macrostates_microstates}

\begin{figure}[!ht]
    \centering    \includegraphics[width=0.9\textwidth]{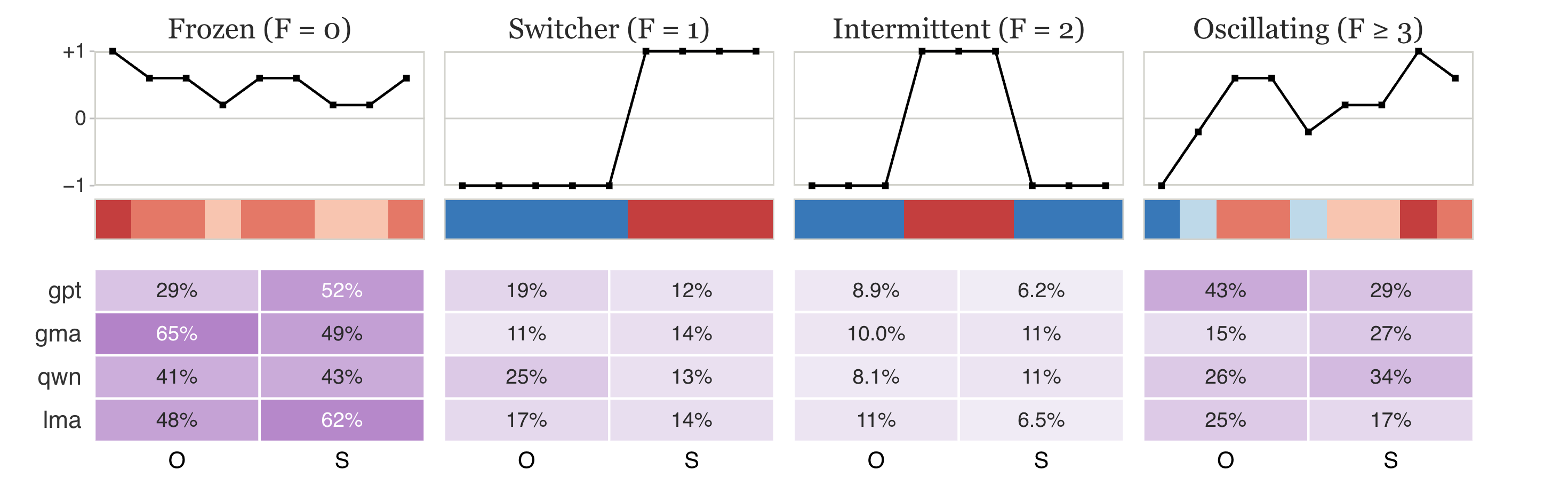}
    \caption{\textit{Individual Archetypes. Individual trajectories are classified into four distinct archetypes, with the prevalence of archetypes varying across models and between objective and subjective questions.} The line plot and the heatmap show individual trajectories $
(\bar{o}_i(0),  \bar{o}_i(1), \dots, \bar{o}_i(T))
$. The table below shows the distribution of individual archetypes across different models (GPT-4o-mini (gpt), Gemma-3n-E4B-it (gma), Qwen3.5-9B (qwn), and Llama-3.1-8B-Instruct (lma)) for the objective (O) and subjective (S) questions.}
\label{fig:microstates}
\end{figure}
\begin{figure}[!ht]
    \centering
    \includegraphics[width=0.99\textwidth]{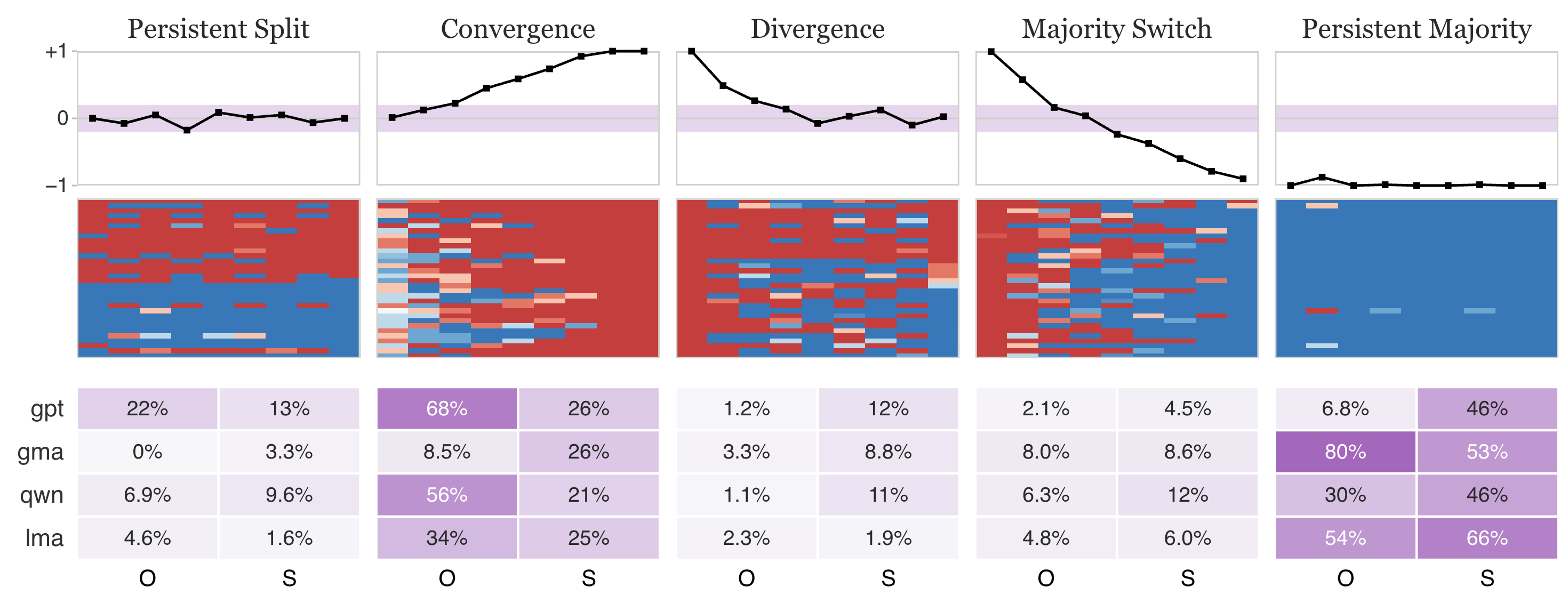}
\caption{\textit{Group Archetypes. Group trajectories exhibit five distinct archetypes whose prevalence varies substantially across models and question types. While some groups uninterestingly lock into a persistent majority, many others converge, diverge, switch majorities, or remain  as a persistently split.} The heatmap and line plot demonstrate an example group trajectory belonging to each of the five group archetypes. Row $i$ of the heatmap shows the individual trajectory, $ (\bar{o}_i(0),  \bar{o}_i(1), \dots, \bar{o}_i(T))
$, of $i$th agent in the group. Column $j$ of the heatmap is a snapshot of the individual opinions at timestep $j$: $ (\bar{o}_1(j),  \bar{o}_2(j), \dots, \bar{o}_N(j))
$. The line plot shows the trend in net opinion \(
n(t)
\) across timesteps. The table below shows the distribution of group archetypes. O and S indicate objective and subjective questions.}\label{fig:macrostates}
\end{figure}

\textit{Individuals.} We denote the opinion of a single agent at a timestep  with $\bar{o}_i(t) \in [ -1, +1]$. An individual trajectory is the sequence of one agent's opinions over time: 
$
(\bar{o}_i(0),  \bar{o}_i(1), \dots, \bar{o}_i(T))
$. By tracing individual trajectories across time, we assign each individual to one of four mutually exclusive and exhaustive archetypes based on how many times they switch sides, \textit{i.e.}, 
\[
F_i=\sum_{t=0}^{T-1}\mathbf{1}\!\left\{\text{sign}(\bar{o}_i(t))\neq \text{sign}(\bar{o}_i(t+1))\right\}=\sum_{t=0}^{T-1}\mathbf{1}\!\left\{s_i(t)\neq s_i(t+1)\right\},\]
where \(\mathbf{1}\{\cdot\}\) is the indicator function. In Figure~\ref{fig:microstates}, we show the (1) \emph{Frozen} agents do not change their opinion throughout the interaction and opinion update rounds ($F_i=0$ switches), (2) \emph{Switchers} start with an opinion and end up with the opposite opinion at the end of the 8 rounds after a single switch ($F_i=1$ switch), (3) \emph{Intermittent} agents are those that switch their opinion and switch it back ($F_i=2$ switches), and (4) \emph{Oscillators} are those that switch their opinion more than twice ($F_i>2$ switches). We observe that the distribution across these archetypes is uneven: frozen agents dominate in most settings followed by oscillators, while switchers and intermittent are comparatively rare.

\textit{Groups.} We refer to the collective opinion trajectory of all $N$ agents as a \textit{group trajectory}, which is an $N \times (T+1)$ matrix whose columns are the successive group opinions and whose rows are the individual trajectories. Running the system on a given $(\{p_i\}_{i=1}^N,\ q,\ J)$ produces one such trajectory, capturing how opinions evolve over time. We also classify group trajectories based on how many times they switch sides. We define a group’s \textit{net opinion} as \(n(t) = \frac{1}{N}\sum_i \bar{o}_i(t)\). Although it is possible to count the number of times \(n(t)\) changes sign, analogous to the classification rule we followed in the individual trajectories, this is not particularly informative for group opinions because minor fluctuations around zero can result in multiple switches that are not meaningful. To address this issue, we define a \textit{split band} \([-\delta, \delta]\) around zero and let
\[
\tilde{n}(t) =
\begin{cases}
+1, & \text{if } n(t) \geq \delta, \\
-1, & \text{if } n(t) \leq -\delta, \\
0,  & \text{otherwise}.
\end{cases}
\]
Using the initial ($\tilde{n}(0)$) and final ($\tilde{n}(T)$) net opinions,\footnote{We use $T=8$ in our experiments as  communities typically converge before 8 iterations.} we classify the group trajectories into $5$ mutually exclusive and exhaustive \textit{group archetypes}.\footnote{Note that the split band is already implicitly defined at the level of individual trajectories. Since
$\bar{o}_i(t)$ is the mean of $K=5$ binary votes in $\{-1,+1\}$, it takes values
only on the grid $\{-1,-0.6,-0.2,+0.2,+0.6,+1\}$, whose smallest nonzero
magnitude is $1/K = 0.2$; and because $K$ is odd, $\bar{o}_i(t)=0$ cannot occur.
The band $(-\delta,\delta)$ with $\delta = 1/K = 0.2$ is therefore already implicitly used
for individual trajectories. We therefore fix the value of $\delta =0.2$, which is the resolution of a single agent's opinion, so
$|n(t)| < 1/K$ means the population average falls below the finest distinction
any individual agent is able to express.} For groups with no clear initial majority (start inside the split band with ($\tilde{n}(0) = 0$), \emph{Persistent Split} ends the 8th round inside the split band ($\tilde{n}(8) = 0$), while \emph{Convergence} ends outside the split band, having converged to a clear majority ($\tilde{n}(8) \neq 0$). For groups with a clear initial majority (start outside the split band with $\tilde{n}(0) \neq 0$), \emph{Persistent Majority} ends the 8th round as the same initial majority opinion ($\tilde{n}(0) \cdot \tilde{n}(8)  > 0$), \emph{Divergence} ends inside the split band ($\tilde{n}(0) \cdot \tilde{n}(8)  = 0$), and \emph{Majority Switch} ends on the opposite side of the split band ($\tilde{n}(0) \cdot \tilde{n}(8)  < 0$).

In Figure \ref{fig:macrostates}, we examine the distribution of different group archetypes and observe that the collective behavior is rich. Notably, \emph{Divergence} and \emph{Majority Switch}, in which the initial majority weakens or is overturned are not rare, reaching $11-12\%$ in GPT-4o-mini and Qwen3.5-9B. Additionally, we also present and discuss $4$ rare non-monotonic group archetypes in Figure \ref{fig:rare_macrostates} and Appendix \ref{apdx:rare_macrostates}. Notably, communication networks can cause consistent differences in group trajectories of the same question (Appendix \ref{apdx:communication_effect}).

\begin{figure}[t]
    \centering
    \includegraphics[width=0.99\textwidth]{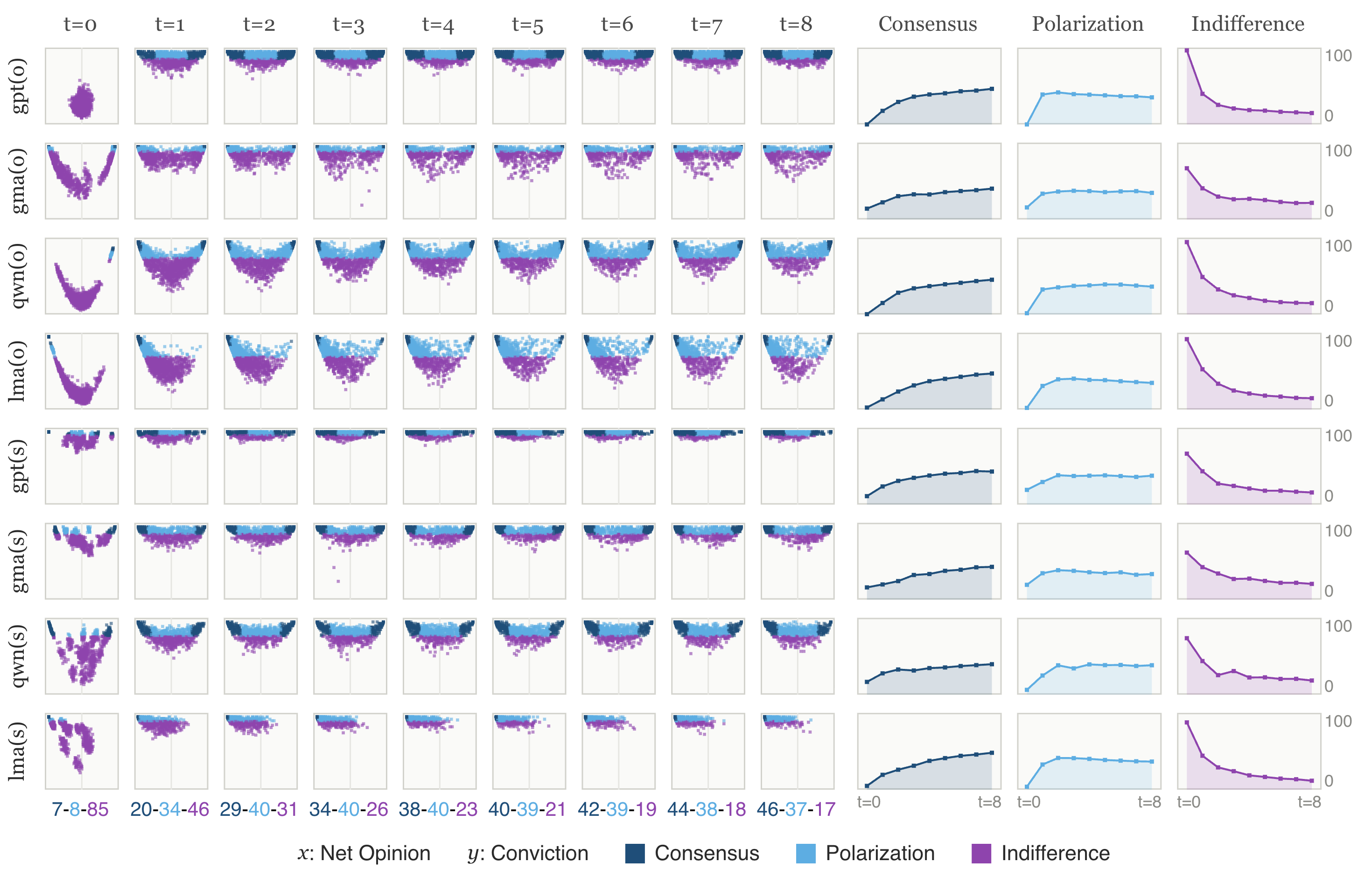}
    \caption{\textit{Conviction Buildup and Consensus Formation.} Each dot represents a group trajectory represented as a point on the net opinion ($x$) and conviction ($y$) plane. By construction, each row is split into approximately equal number of indifference, consensus and polarization examples. The numbers below each column indicate the percentage of examples that fall into each characteristic regimes at that time step, aggregated across all model–regime pairs.}
    \label{fig:3_Observations_modes}
\end{figure}

\begin{figure}[t]
    \centering
    \includegraphics[width=0.99\textwidth]{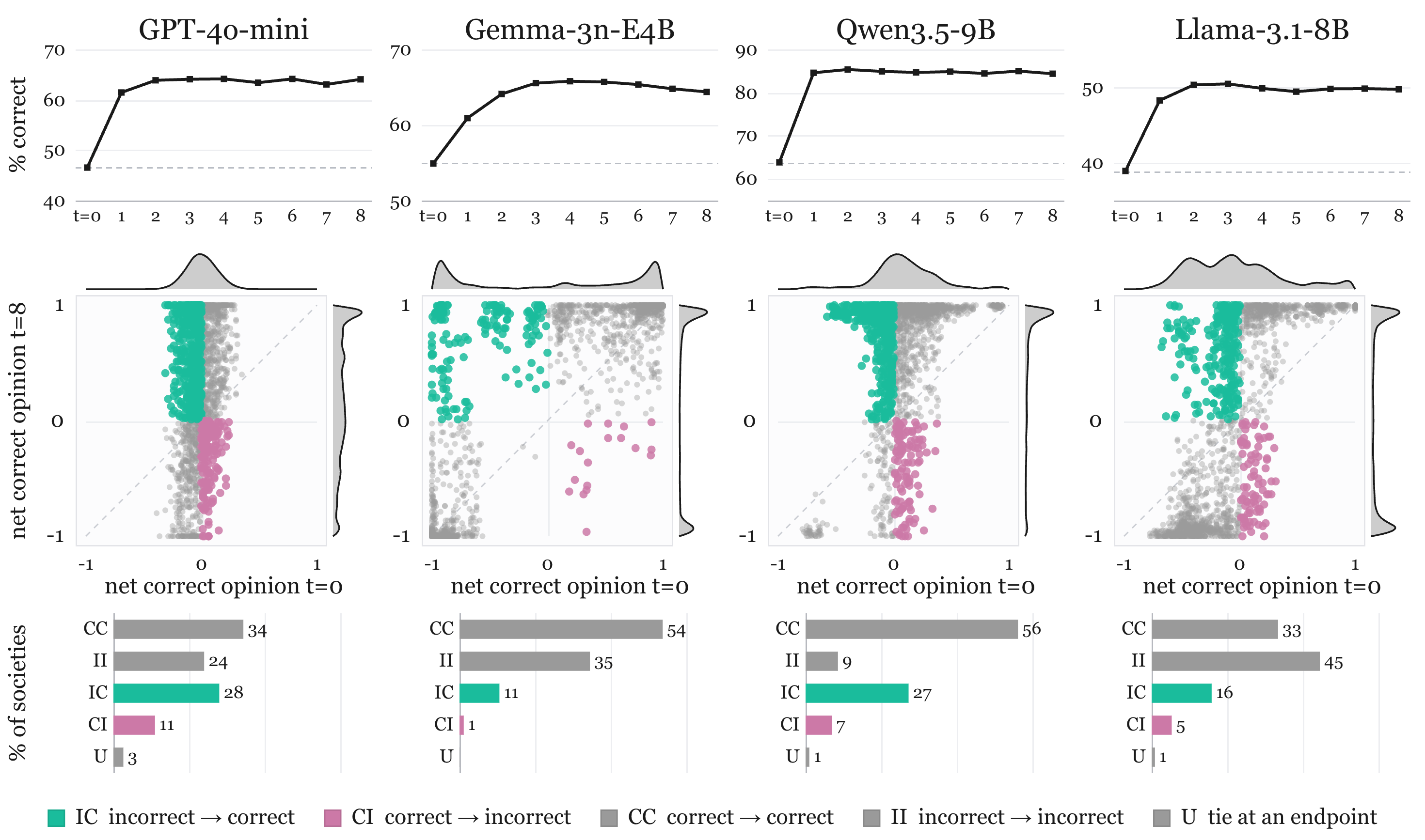}
    \caption{\textit{Truth-seeking tendency.} How opinions move toward the correct answer over a run ($6,400$ group trajectories from objective questions only, from signed random graphs and the square and triangular lattices). In a group trajectory, an opinion is represented as true if net opinion has the correct sign. \emph{Row 1:} $y$-axis: percentage of group trajectories, where the net opinion $n(t)$ has the correct sign at round $t$. $x$-axis: timesteps. \emph{Row 2:} Each point represents a group trajectory, which is plotted based on its initial ($t=0$, $x$-axis) and final ($t=8$, $y$-axis) net opinion $n(t)$ multiplied by the sign of the correct answer of the questions to get \textit{net correct opinion}: $n(t) \cdot y_{gt}$, where $y_{gt} \in \{-1, +1\}$ denotes the ground truth answer to the objective question. \emph{Row 3:} Groups switch between being correct and incorrect at $t=0$ and $t=8$. CC (correct$\to$correct), CI (correct$\to$incorrect), IC (incorrect$\to$correct), II (incorrect$\to$incorrect), and U (undecided, $50\text{-}50$ tie at $t=0$ or $t=8$). The two transition categories CI and IC are highlighted.}
\label{fig:3_Observations_ts}
\end{figure}

\subsection{Conviction Buildup and Consensus Formation}
\label{sec:obs_phase_char}

When the net opinion, \(
n(t) = \frac{1}{N}\sum_i \bar{o}_i(t)
\), is close to $+1$ or $-1$, the collective opinion is straightforward to interpret as it means most individuals hold strong opinions that are aligned in the same direction (\textit{consensus}). In contrast, a net opinion close to zero can arise in two fundamentally different ways. In the first case (\textit{polarization}), individuals hold strong opinions (close to +1 or -1), but they are evenly split between opposing viewpoints.  In the second case (\textit{indifference}), most individuals are uncertain or indifferent about their position, and aggregating these weak preferences naturally yields a collective opinion near zero. To characterize this distinction, we define conviction as a measure of how strongly opinionated agents are at a given timestep \(
c(t) = \frac{1}{N}\sum_i \bar{o}_i^2(t)
\). 

In Figure~\ref{fig:3_Observations_modes}, we follow the joint distribution of net opinion ($x$) and conviction ($y$) for all eight model-regime settings as the interaction unfolds from $t=0$ (left) to $t=8$ (right), the trailing columns count the communities in each regime at every round. Conviction rises over time in all settings. We define three characteristic regimes: consensus (high $|n(t)|$, high $c(t)$), polarization (near-zero $|n(t)|$, high $c(t)$), and indifference (near-zero $|n(t)|$, near-zero $c(t)$). We select conviction and net opinion  thresholds such that an equal number of points belong to each of the three characteristic regimes in each row.\footnote{The selected thresholds are reported in Table \ref{tab:mode_thresholds} from Appendix \ref{apdx:mode_thresholds}.} We observe that early timesteps are dominated by indifference, and as communities progressively transition toward consensus or polarization, indifference decreases while consensus increases monotonically.

\subsection{Truth Seeking Tendency}\label{sec:truth_seeking}

As the group trajectories reach consensus, it is important to know whether this consensus converges toward the correct answer on questions for which ground-truth answers  are available.

At a given timestep, we read off the community's answer from the sign of the net opinion $n(t)$. In Figure~\ref{fig:3_Observations_ts}, we observe that communities tend to move toward the correct answers for objective questions. The percentage of communities whose $n(0)$ prior to any interaction is the correct answer is given by the $t=0$ point in the line plot. As the agents exchange messages, we observe that the majority vote of the community has a tendency to move towards the correct answer, with the strong improvements observed for Gemma, GPT, and Qwen.

\textit{Majority Switches.} The second row in Figure~\ref{fig:3_Observations_ts} shows the communities that initially give incorrect answers and later switch to correct answers, as well as communities that initially give correct answers and later switch to incorrect answers. We observe that, for all models, switches from incorrect to correct are more common than switches in the opposite direction, which is consistent with our truth-seeking findings.

We also note that language-model performance often improves with additional inference-time compute, as demonstrated by Mixture-of-Agents and multi-agent debate frameworks that iteratively aggregate, critique, and refine responses \citep{wang2024mixtureofagentsenhanceslargelanguage,du2023improvingfactualityreasoninglanguage,liang-etal-2024-encouraging}. In our setting, networks of language-model agents likely benefit from a similar mechanism.

On subjective questions, there is not a well-defined notion of correct or incorrect answer. Instead, we investigated whether the collective opinion of the group tends to drift to the left or right of the political spectrum. Interestingly, it's more likely for the majority opinion of the agents to shift from left to right over time than the other direction. We discuss this further in Appendix \ref{apdx:political_lean}.

\section{Statistical Mechanics Model of Agents' Behavior}\label{sec:method}

We model community as a system in which individuals adjust their opinions in response to social pressure. We define $s_i$ as the opinion state of agent $i$ and the interaction coefficient $J_{ij}$ as the influence of agent $j$ on agent $i$. Then, two key modeling assumptions give rise to a mathematically tractable model of multi-agent interactions and opinion forming. First, we assume that individuals would like to minimize the social pressure. Consider the term $a_i=\sum_j J_{ij}s_j$, which represents the perceived population opinion that an individual $i$ faces from their social environment. Mathematically, $a_i s_i > 0$ when, on average, agent $i$ sits on the same side as its friendly connections and on the opposite side of its unfriendly connections, and $a_i s_i < 0$ otherwise. Using this observation, we define the term $-\sum_i a_i s_i =-\sum_i \sum_j s_i J_{ij}s_j$ as the total social pressure in the community. When there is a misalignment, the individual experiences are pressured to conform or adjust. Second, for a given question, we assume that each individual has a tendency to hold an opinion of their own, say $\tilde s_i$. When $\tilde s_i$ and $s_i$ disagrees, this creates yet another pressure point. Thus, we define the term $-\lambda \sum_{i} \tilde s_i s_i$ as the personal pressure, where $\lambda\geq 0$ is a parameter quantifying the relative importance of the two contributions. Our hypothesis is that the population opinions, $s = (s_1, s_2, ... , s_N)$, favor lower-pressure configurations and stochastically evolve toward configurations that minimize this total pressure.

Consequently, with $g_i=\lambda \tilde s_i$, we can define a loss, or equivalently, an energy function as
$$
E(s) = -\frac{1}{2}\sum_i \sum_j s_i J_{ij}s_j - \lambda \sum_{i} \tilde s_i s_i \;=\; -\frac{1}{2} \sum_{i=1}^N \sum_{j=1}^N J_{ij}\,s_i s_j \;-\; \sum_{i=1}^N g_i\, s_i .
$$
The symmetric coupling coefficient $J_{ij} \in \{+1,0,-1\}$ quantifies the nature of social influence between agents $i$ and $j$, while the intrinsic field $g_i \in \mathbb R$ captures agent $i$'s predisposition on the issue. The first term rewards configurations that are consistent with the underlying social relationships (assumption 1), whereas the second term biases opinions toward agents' individual tendencies (assumption 2). In an ideal world, each individual would hold an opinion aligned with their intrinsic predisposition while also satisfying all interpersonal constraints. 

Fixing all other agents' opinions, the local field the agent $i$ experiences can be expressed as
\(f_i = \sum_{j\neq i} J_{ij}\,s_j + g_i
\). We can derive (see Appendix \ref{apdx:discrete-updates})  the probability of an agent having opinion $s_i = +1$ to be
\[
P(s_i=+1\mid s_{-i})
= \sigma\!\left(h_i\right), \quad \text{where} \quad h_i = \beta \sum_{j} J_{ij}\,s_j(t) + g_i
\]
where $s_{-i}$ is the opinion state of all other agents. This shows that an agent's next opinion is a logistic function of (a) the peer pressure $\sum_j J_{ij}s_j(t)$ flowing in along the signed graph, scaled by the coupling $\beta$, and (b) an intrinsic field $g_i$ encoding the leaning of the persona of agent $i$ for this question in the absence of any peer influence.

\textit{Three Couplings.}\label{sec:dissecting-couplings} We observe that, in a system of interacting language models, (a) the existence of a connection itself creates a pull and (b)  the pull and push strengths of discordant and concordant connections differ. To account for these cases, we introduce a three coupling model. 
\[
P\big(s_i = +1\big) = \sigma\!\left(\beta^{+} \sum_{j} J^{+}_{ij}\,s_j(t) + \beta^{-} \sum_{j} J^{-}_{ij}\,s_j(t) + \beta_{0} \sum_{j} |J_{ij}|\,s_j(t) + g_i\right).
\]
where $\beta_{0}$ measures the effect of having a connection, $\beta^{+}$ measures the effect of having a concordant connection, and $\beta^{-}$ measures the effect of having a discordant connection. The total weight an agent places on a friendly neighbor is then $\beta^{+} + \beta_{0}$ and on an unfriendly neighbor $\beta_{0} - \beta^{-}$. Note that this is equivalent to using only two couplings ( $\beta^{+}$ and  $\beta^{-}$), but we retain this version because it makes comparisons to the single coupling case explicit. Further, due to a two-stage fitting process described below, this choice allows a direct interpretation of the terms $\beta^{\pm}$ by decoupling the effect of being a neighbor from the valence of the connection.

\paragraph{Fitting.} We parameterize $g_i$ as a function of embeddings, $ g_i = w^\top \phi_i$, where $\phi_i$ concatenates the agent's persona embedding, the question embedding, and their interaction. We fit the $\beta$s and $w$ by running gradient-descent to minimize the cross-entropy between each agent's predicted next opinion and its observed value on the one-step transitions. For the three-coupling model, fitting proceeds in two stages: we first estimate \(\beta_0\), then, holding \(\beta_0\) fixed, estimate \(\beta^+\) and \(\beta^-\) jointly. Appendix~\ref{apdx:fitting} provides the details of the objective, optimization, and the train--test splits. 

\paragraph{Continuous-time extension.} In Appendix \ref{apdx:async_update}, we relax the assumption that every agent revises their opinion at discrete time steps and instead let each agent reconsider their opinion at independent random times. In  Appendix~\ref{apdx:continuous-extension}, we model this relaxation via a continuous-time extension of our model with a mean-field ODE derived from the master equation that captures probability flux between states. This mean-field ODE contains a parameter $\varepsilon \in (0,1)$ representing the rate of agent updates.  %
We present this continuous-time variant alongside the discrete rules and discuss its advantages in Appendix~\ref{apdx:res_utility_of_continuous} and Appendix \ref{apdx:async_update}.

\paragraph{Baselines.} We compare against Persistence, Interaction Free, and Mean-Field baselines. \textit{Persistence} predicts \(\hat{s}_i(t+1)=s_i(t)\) in one step predictions and $\hat{s}_i(t)=s_i(0)$ in rollouts. \textit{Interaction-Free} keeps personal pressure
but removes all social-pressure terms, and \textit{Mean-Field} retains both personal and
social pressure but ignores graph structure by encoding social pressure as pulling each agent
toward the global average opinion. This tests whether explicit signed-network structure adds predictive value beyond a global consensus signal. Details of the baselines are explained in Appendix \ref{apdx:baselines}.

\section{Applications of the Statistical Physics Model}\label{sec:results}
We next demonstrate that our statistical physics model can predict
agent dynamics for unseen graphs and questions, and reproduces the population-level
distribution of collective outcomes when rolled out from observed initial conditions.
Furthermore, this model suggests that the agent communities operate below the critical social temperature,
which explains why interacting communities do not remain indifferent but tend to form consensus or polarize. Which high conviction state they reach is determined by the social ties and the intrinsic field. We observe that the pull of friendly (concordant) edges dominates the push of unfriendly (discordant) edges, which favors consensus over polarization, and the neighbors holding the correct answer pull harder than ones holding the wrong answer, which drives truth seeking.

\subsection{Prediction and Generalization}\label{sec:results_prediction}
\begin{table}[t]
\centering
\small
\begin{adjustbox}{max width=\textwidth}
\begin{tabular}{l cc cc cc cc}
\toprule
 & \multicolumn{2}{c}{GPT-4o-mini} & \multicolumn{2}{c}{Gemma-3n-E4B} & \multicolumn{2}{c}{Qwen3.5-9B} & \multicolumn{2}{c}{Llama-3.1-8B} \\
\cmidrule(lr){2-3}\cmidrule(lr){4-5}\cmidrule(lr){6-7}\cmidrule(lr){8-9}
Method & In-d. & Out-d. & In-d. & Out-d. & In-d. & Out-d. & In-d. & Out-d. \\
\midrule
\multicolumn{9}{l}{\textit{Subjective Questions}} \\
\midrule
Persistence & 50.0 (64.5) & 50.0 (64.0) & 50.0 (59.3) & 50.0 (59.3) & 50.0 (65.1) & 50.0 (65.1) & 50.0 (63.2) & 50.0 (61.7) \\
Interaction-Free & 50.9 (50.9) & 48.9 (48.9) & 60.8 (60.8) & 59.6 (59.6) & 49.1 (49.1) & 47.5 (47.5) & 47.7 (47.7) & 45.6 (45.6) \\
Mean-Field & 57.9 (56.0) & 59.2 (57.2) & 68.2 (64.5) & 67.9 (65.8) & 65.2 (64.4) & 65.6 (65.5) & 54.9 (56.8) & 53.6 (57.4) \\
Discrete Update & 77.6 (69.0) & 75.6 (67.1) & 62.3 (61.0) & 61.3 (59.9) & 64.5 (57.1) & 60.3 (54.4) & 53.9 (49.9) & 51.6 (47.8) \\
\quad + 3 Couplings & \textbf{86.0 (76.2)} & \textbf{86.3 (77.2)} & \textbf{75.4 (67.8)} & \textbf{75.1 (66.8)} & \textbf{81.2 (70.8)} & \textbf{78.8 (69.2)} & \textbf{82.4 (68.4)} & \textbf{80.8 (69.5)} \\
\midrule
\multicolumn{9}{l}{\textit{Objective Questions}} \\
\midrule
Persistence & 50.0 (55.6) & 50.0 (55.0) & 50.0 (61.8) & 50.0 (62.2) & 50.0 (53.9) & 50.0 (53.2) & 50.0 (57.0) & 50.0 (56.8) \\
Interaction-Free & 49.9 (49.9) & 49.3 (49.3) & 59.0 (59.0) & 59.1 (59.1) & 44.6 (44.6) & 45.4 (45.4) & 50.0 (50.0) & 50.0 (50.0) \\
Mean-Field & 65.4 (55.2) & 65.3 (55.5) & 72.6 (67.3) & 72.7 (68.3) & 73.6 (62.1) & 73.4 (61.1) & 64.7 (61.2) & 64.5 (60.5) \\
Discrete Update & 70.1 (61.5) & 67.8 (60.4) & 59.7 (58.9) & 59.7 (59.2) & 50.6 (47.8) & 49.8 (48.3) & 50.9 (50.0) & 50.6 (50.0) \\
\quad + 3 Couplings & \textbf{86.2 (68.2)} & \textbf{85.0 (65.9)} & \textbf{80.5 (68.5)} & \textbf{80.8 (68.4)} & \textbf{86.3 (65.6)} & \textbf{85.2 (64.2)} & \textbf{77.3 (62.8)} & \textbf{76.4 (60.8)} \\
\bottomrule
\end{tabular}
\end{adjustbox}
\caption{\textit{Prediction.} Balanced accuracy of predictions on held-out test questions. Training and test sets each include $320= 10 \times 8 \times 4$ samples for subjective and  $640 = 20 \times 8 \times 4$ for objective questions (questions $\times$ $J$s $\times$ episodes). The numbers represent the accuracy of predicting $s(t+1)$ given $s(t)$ (one-step), and the accuracy of predicting $s(t)$ for $t=1,\dots , T$ given $s(0)$ is reported in parentheses (rollout). Four of the 8 graphs in the test set are graphs seen during training, denoted in-d, and the other four graphs are held-out during training, denoted out-d. 
\label{tab:prediction_models}}
\end{table}

\begin{table}[t]
\centering
\small
\setlength{\tabcolsep}{3pt}
\begin{adjustbox}{max width=\textwidth}
\begin{tabular}{l cccccccccccccccc}
\toprule
 & \multicolumn{8}{c}{Subjective} & \multicolumn{8}{c}{Objective} \\
\cmidrule(lr){2-9}\cmidrule(lr){10-17}
 & \multicolumn{2}{c}{Random} & \multicolumn{2}{c}{Low Rank} & \multicolumn{2}{c}{Square} & \multicolumn{2}{c}{Triangular} & \multicolumn{2}{c}{Random} & \multicolumn{2}{c}{Low Rank} & \multicolumn{2}{c}{Square} & \multicolumn{2}{c}{Triangular} \\
Method & 1-step & Roll. & 1-step & Roll. & 1-step & Roll. & 1-step & Roll. & 1-step & Roll. & 1-step & Roll. & 1-step & Roll. & 1-step & Roll. \\
\midrule
Persistence & 50.0 & 64.0 & 50.0 & 57.0 & 50.0 & 65.2 & 50.0 & 57.4 & 50.0 & 55.0 & 50.0 & 53.1 & 50.0 & 55.1 & 50.0 & 47.0 \\
Interaction-Free & 48.9 & 48.9 & 46.4 & 46.4 & 46.8 & 46.8 & 47.4 & 47.4 & 49.3 & 49.3 & 43.4 & 43.4 & 46.8 & 46.8 & 49.6 & 49.6 \\
Mean-Field & 59.2 & 57.2 & 58.0 & 63.2 & 65.3 & 61.7 & 64.9 & 62.4 & 65.3 & 55.5 & 51.3 & 60.5 & 70.5 & 60.1 & 73.4 & 56.9 \\
\midrule
Discrete Update & 75.6 & 67.1 & 80.1 & 75.6 & 88.4 & 78.3 & 84.3 & \textbf{73.5} & 67.8 & 60.4 & 82.9 & 69.3 & 92.7 & 65.8 & 87.7 & 58.8 \\
\quad + Three Couplings & \textbf{86.3} & \textbf{77.2} & \textbf{95.1} & \textbf{89.3} & \textbf{92.6} & \textbf{83.5} & \textbf{85.3} & 73.2 & \textbf{85.0} & \textbf{65.9} & \textbf{97.8} & \textbf{80.7} & \textbf{94.7} & \textbf{70.9} & \textbf{88.1} & \textbf{59.7} \\
\bottomrule
\end{tabular}
\end{adjustbox}
\caption{\textit{Generalization.} We report the generalization to unseen graph families and report the balanced accuracy of GPT-4o-mini's update rules on unseen random graphs,  the low-rank graphs, and the square / triangular lattices. We report both one-step and rollout accuracies as described in Table \ref{tab:prediction_models}.\label{tab:generalization}}
\end{table}

\textit{Metric.} Since opinions change rarely ($s_i(t) \approx s_i(t+1)$) and the two labels ($+1$ and $-1$) are not equally common, plain accuracy rewards a rule that copies the current opinion or one that always predicts the more common label. We score predictions with \emph{balanced accuracy} where we balance for the flips and class imbalance in the dataset. We classify the observed transitions into four groups, by whether the agent's next opinion state flips or stays ($s_i(t{+}1) \neq s_i(t)$ or $s_i(t{+}1) = s_i(t)$) and by the value of that next opinion state ($+1$ or $-1$). Our balanced accuracy metric is the unweighted mean of the accuracies on these four groups (see Appendix~\ref{apdx:metric}).

\textit{Prediction.} We observe that the discrete update with three couplings is the best method in every column of Table~\ref{tab:prediction_models}: $75$--$86$ one-step (predicting $s(t+1)$ given $s(t)$) balanced accuracy across the four models and both question types, and $61$--$77$ in rollouts (predicting $s(t)$ for $t=1,\dots , T$ given $s(0)$). Splitting the interaction parameter $\beta$ matters since with a single $\beta$, the same rule falls close to random chance in some instances  ($53.9$ with Llama-3.1-8B-Instruct, subjective and $50.6$ with Qwen3.5-9B, objective). The mean-field baseline performs in the mid-$50$s (GPT-4o-mini and Llama-3.1-8B-Instruct, subjective) to the low-$70$s (objective one-step). The interaction-free baseline is close to chance for three of the four models. For Gemma-3n-E4B it reaches $59$--$61$, indicating that some of its agents' opinions are predictable from persona and question alone, without considering any interactions.

\textit{Generalization.} For the three-coupling rule, the in-distribution and out-of-distribution columns of Table~\ref{tab:prediction_models} differ by at most $2.4$ points; the largest gap for any method is $4.2$. In Table~\ref{tab:generalization}, we test the model further by evaluating it on three graph families not included in the training set. In addition to the held-out random graphs, we consider low-rank social graphs, square lattices, and triangular lattices. The three-coupling rule is the best method in $15$ of the $16$ columns, with one-step balanced accuracy between $85.0$ and $97.8$. Its rollout performance declines more ($59.7$--$89.3$), but remains above every baseline. The low-rank and square lattice families appear easier than the random graphs used for fitting, possibly because of differences in edge count or the proportion of negative edges. The triangular lattice, on the other hand, has comparable one-step accuracy, but its rollout accuracy is lower than that of random graphs ($73.2$ vs $77.2$ subjective, $59.7$ vs $65.9$ objective).

\textit{Larger communities with heterogeneous frontier models}. We conducted additional experiments to demonstrate that our statistical physics model can accurate predict the collective dynamics of communities with larger number of agents consisting of different frontier LLMs (\ref{apdx:frontier_mixed}). 
 We construct community of mixed-family group of $ 64$ agents consisting of 32 agents powered by GPT-5.6-sol and 32 agents powered by DeepSeek-V4-Flash.
 The discrete three-coupling rule fit on the training questions remains the best method on held-out questions under balanced accuracy, reaching 81.9 (70.4 rollout) on subjective and 80.0 (67.4) on objective questions over all agents \ref{tab:frontier_balanced}. The model predicted similarly well for both LLM model families. This result shows that our model can also predict the dynamics of larger communities made up of heterogeneous frontier models.

\begin{figure}[t]
    \centering
    \includegraphics[width=0.99\textwidth]{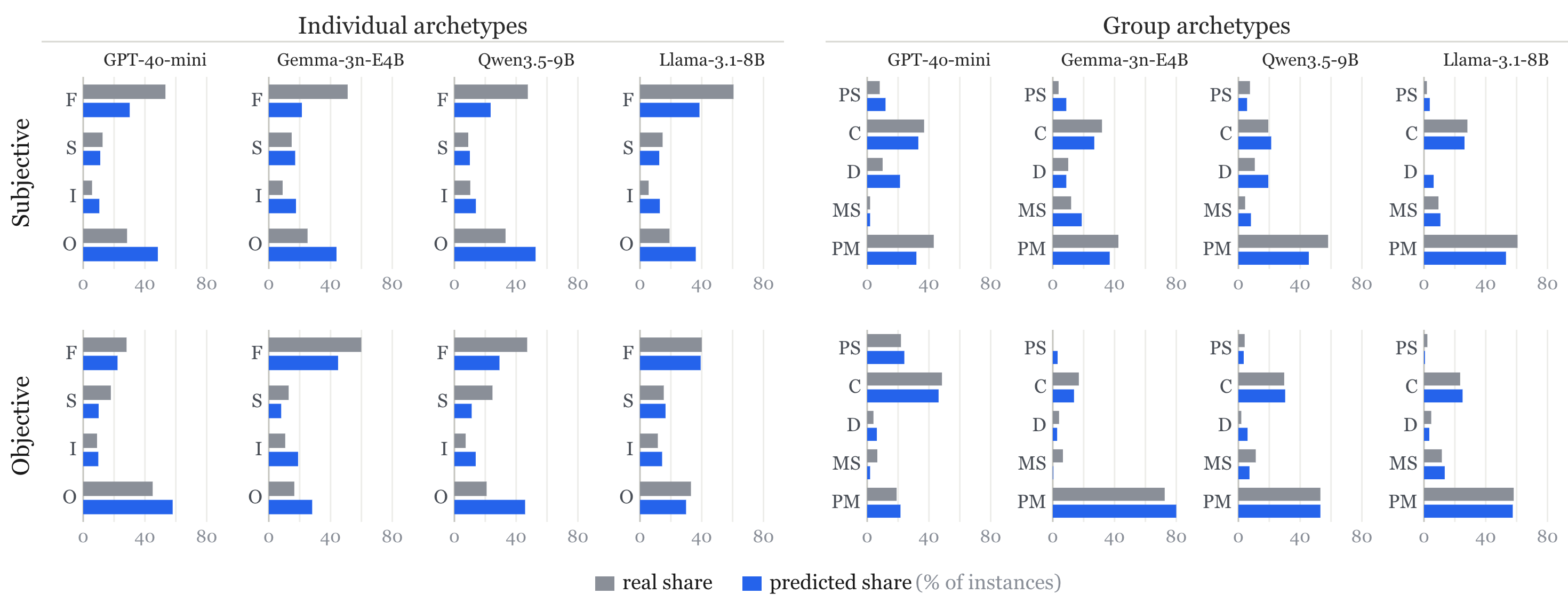}
\caption{\textit{Distribution of Individual and Group Archetypes.} Shares
of individual and group archetypes are shown as horizontal bars (F~(Frozen), S~(Switcher), I~(Intermittent), O~(Oscillating); PS~(Persistent Split), C~(Convergence), D~(Divergence), MS~(Majority Switch), PM~(Persistent Majority)). The shares are computed on the held-out test set questions using the in-distribution random graphs by running stochastic rollout of the fitted Discrete Three couplings model from the real $s(0)$ for $8$ rounds, in which each agent's next opinion is sampled as $s_i(t{+}1)=+1$ with probability $\sigma(h_i(t))$ rather than thresholded at at probability $1/2$. \label{fig:macrostate_distribution}}
\end{figure}

\subsection{Distributions of Archetypes}\label{sec:results_distributions}

\textit{Predicting the Distribution of Group Archetypes.} We also investigate whether the fitted model, run from the real initial opinions $s(0)$ as a roll-out, produces communities that look like the real ones in aggregate. Because the fitted rule is a probability rather than a decision, we sample each agent's next opinion from $P(s_i(t{+}1)=+1)=\sigma(h_i(t))$, where $h_i(t)$ is the fitted logistic argument of Section~\ref{sec:method}, instead of thresholding it at $\sigma = 1/2$, so that the rollouts carry the noise the rule assigns. Figure~\ref{fig:macrostate_distribution} compares the shares of the individual and group archetypes of Section~\ref{sec:macrostates_microstates} between the held-out communities and these rollouts (for details on the roll-outs, see Appendix~\ref{apdx:rollouts}).

\textit{Individual and Group Archetypes.} In Figure~\ref{fig:macrostate_distribution}, the group archetype shares deviate by only $\sim\!3$ points for objective questions and $\sim\!5$ points for subjective questions (mean absolute deviation). The largest single gap is $\sim\!10$ points (Qwen3.5-9B subjective, Persistent Majority) and the two dominant classes correct in every cell. Meanwhile, the model inflates the oscillator count in individual archetypes, but this appears to largely cancel out in the group archetype distributions. The fitted dynamics therefore reproduce the collective statistics even when they misstate how often an individual agent changes its mind.  We note that this observation demonstrates that the fitted dynamics produce the right mix of outcomes, not that they place a given group in the right archetype.

\textit{Reproducibility of Group Trajectories.} We observe that group trajectories are only partly predictable. We run each group as four episodes with identical personas, graph and question that differ only in sampling randomness. We compare how often one episode's group archetype matches the other samples taken with the identical personas, graph, and question. Figure ~\ref{fig:predictability_macro_subjective_test_J03_cv_row} in Appendix~\ref{apdx:replica_predictability} reports this ceiling. Episodes frequently disagree on the group archetype, which is why we evaluate the fitted rule at the level of the distribution.

\subsection{Critical Temperature and Conviction Buildup}\label{sec:results_temperature}

\begin{figure}[t]
    \centering
    \includegraphics[width=0.99\textwidth]{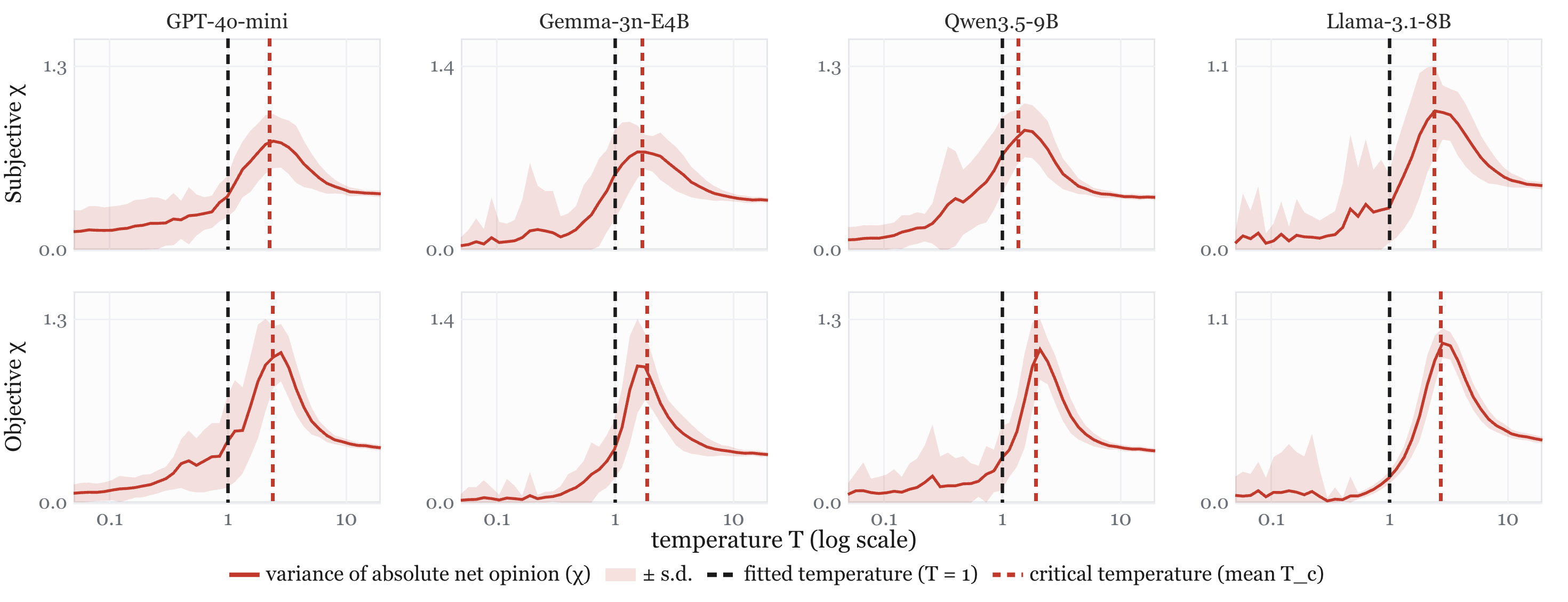}
\caption{\textit{Critical temperature.} 
Variance of absolute net opinion $\chi$, whose peak gives the critical temperature $\mathcal{T}_c$ for each community. Individual communities are reported in Figure \ref{fig:agent_model_tc_mean_test_J03_fitted_nomf_tcmean_qea} from  Appendix \ref{apdx:per_society_transitions}. Here we report instead the mean of the community critical temperatures $T_c$. 
The bold black dashed line is the fitted operating point $\mathcal{T}=1$ of the three coupling model. Every question from the held-out test questions evaluated on the $4$ in-distribution random test graphs (each of the $4$
graphs of the family $\times$ the questions, $20$ objective / $10$ subjective
questions).\label{fig:agent_model_tc_mean_test_J03_fitted_nomf_tcmean}}
\end{figure}

In Section \ref{sec:obs_phase_char}, we saw that conviction goes up over rounds and communities end up mostly in consensus or polarization. In statistical mechanics, whether a system stays ordered depends on where its temperature stands compared to the critical temperature. In this section, we introduce a temperature parameter into our model and check where our fitted communities sit relative to the critical temperature.
In Figure \ref{fig:agent_model_tc_mean_test_J03_fitted_nomf_tcmean}, we reintroduce a temperature parameter to examine our model’s behavior under different temperatures. The fitted-model rollouts use temperature $\mathcal{T}=1$. We explore 41 log-spaced temperature values from 0.05 to 20. We initialize each rollout using the community’s initial opinions at $t=0$ and apply the fitted rule for $500$ steps. Details of the experiment are discussed in Appendix \ref{apdx:per_society_transitions}.

\textit{Variance of Absolute Net Opinion.} We plot the  variance of the absolute net opinion scaled by the number of agents, $\chi$ for different temperatures $\mathcal{T}$.\footnote{In this experiment, we set $K=1$, hence, $o_i(t) = \bar{o}_i(t) = s_i(t)$ and $n(t) = \frac{1}{N}\sum_i \bar{o}_i(t) = \frac{1}{N}\sum_i s_i(t)$.}
$$\chi \;=\; N\,\mathrm{Var}_t\big(|n(t)|\big)$$
It records how much the net opinion moves throughout the rollout. At high $\mathcal{T}$ each agent follows noise, so the net opinion stays near zero at every step and barely moves. At low $\mathcal{T}$ the community is locked into one arrangement, so the net opinion is large but again barely moves. $\chi$ is small in both cases and large only in between.

\textit{Critical Temperature.} We take each community's critical temperature $\mathcal{T}_c$ to be the temperature at which $\chi$ peaks. This peak marks a transition: it is the temperature at which the community is undecided between the two behaviors described above. On either side, the net opinion is stable, either near zero (indifference) or near its ordered value by the couplings (consensus or polarization). At the critical temperature, it can swing between them. This critical temperature marks a finite-size analogue of a phase transition between the high conviction regimes (polarization and consensus) and the low conviction regime (indifference).\footnote{A phase transition in physical terms exists when $N \rightarrow \infty$.}   

\textit{Conviction Build-up.} Figure~\ref{fig:agent_model_tc_mean_test_J03_fitted_nomf_tcmean} shows that the fitted operating point lies below the critical temperature in every dataset--model cell. Below $\mathcal{T}_c$, the indifferent states are unstable. One way to understand this is that aligned neighbors raise an agent's local field, so an agent that agrees with its neighbors is unlikely to move out of its current position when there is low stochasticity. Thus, the model predicts that conviction grows over rounds rather than fluctuating, as observed in Section~\ref{sec:obs_phase_char}.

\subsection{Consensus Formation and Truth Seeking}\label{sec:results_truthseeking}

\begin{table}[t]
\centering
\small
\setlength{\tabcolsep}{4pt}
\begin{tabular}{l ccc ccc ccccc}
\toprule
& \multicolumn{3}{c}{Three couplings (O)}
& \multicolumn{3}{c}{Three couplings (S)}
& \multicolumn{5}{c}{Five couplings (O)} \\
\cmidrule(lr){2-4} \cmidrule(lr){5-7} \cmidrule(lr){8-12}
Model & $\beta^{+}$ & $\beta^{-}$ & $\beta_{0}$
      & $\beta^{+}$ & $\beta^{-}$ & $\beta_{0}$
      & $\beta^{+}_{T}$ & $\beta^{+}_{F}$ & $\beta^{-}_{T}$ & $\beta^{-}_{F}$ & $\beta_{0}$ \\
\midrule
GPT-4o-mini   & $+2.10$ & $+0.75$ & $+0.93$ & $+2.39$ & $+1.08$ & $+0.61$ & $+2.29$ & $+1.98$ & $+0.70$ & $+0.85$ & $+0.93$ \\
Gemma-3n-E4B  & $+0.29$ & $+0.23$ & $+0.96$ & $+0.44$ & $+0.35$ & $+0.55$ & $+0.34$ & $+0.20$ & $+0.17$ & $+0.36$ & $+0.96$ \\
Qwen3.5-9B    & $+0.90$ & $+0.50$ & $+1.01$ & $+1.12$ & $+0.65$ & $+0.64$ & $+1.03$ & $+0.77$ & $+0.40$ & $+0.80$ & $+1.01$ \\
Llama-3.1-8B    & $+0.97$ & $+0.61$ & $+0.97$ & $+1.35$ & $+0.67$ & $+0.99$ & $+1.15$ & $+0.84$ & $+0.59$ & $+0.61$ & $+0.97$ \\
\bottomrule
\end{tabular}
\caption{Fitted coefficients of the three- and five-coupling discrete-time update rules. Because the unsigned drive lies in the span of the signed drives, the individual coefficients are not jointly identifiable; we therefore fix the decomposition by a two-stage protocol: $\beta_{0}$ is first fitted alone (a sign-agnostic model with no signed couplings), then held fixed while the signed couplings are refitted. Stage one is identical for both rules, so within a dataset the three- and five-coupling rows share $\beta_{0}$. The identifiable combinations ($\beta^{+}+\beta_{0}$, $\beta_{0}-\beta^{-}$, and the truth gaps $\beta^{+}_{T}-\beta^{+}_{F}$, $\beta^{-}_{T}-\beta^{-}_{F}$) agree with the jointly fitted (ridge-resolved) values to two decimals. The five-coupling rule splits each drive by whether the pull is toward the correct answer, so it is defined for the objective dataset only.\label{tab:couplings_five}}
\end{table}

\textit{Consensus Formation.} Below the critical social temperature the community tends to settle into an ordered (high conviction) state over the long run. But, which ordered state it reaches depends on the strength of the personal and social pressures. Focusing on the social term, a stable split can only sustain itself if discordant connections are strong enough to hold two groups apart. All four models show the same pattern in Table~\ref{tab:couplings_five}, with $\beta^{+}$ consistently larger than $\beta^{-}$. The effective coefficient on concordant edges ($\beta^{+}+\beta_{0}$) lies between $0.99$ and $3.03$ across all models and question types, while the effective coefficient on discordant edges ($\beta_{0}-\beta^{-}$) never exceeds $0.73$ and is negative or negligible in two cases ($-0.47$ for GPT-4o-mini and $-0.01$ for Qwen3.5-9B, subjective). A stable split requires repulsion from discordant ties to be comparable in strength to the attraction from concordant ties. However, we observe that the discordant ties are not as strong as the concordant ones and, hence, are too weak to hold two camps apart. Since the fitted dynamics operate below the critical temperature with strong concordant and weak discordant couplings, social pressures favor consensus, consistent with the trend observed in Section~\ref{sec:obs_phase_char}. \footnote{We note that strong personal pressures can nonetheless sustain polarization when agents are intrinsically motivated to take different sides (oppositely signed strong $g_i$).}

\textit{Truth Seeking.} To understand the truth seeking tendency of the system, we investigate whether a neighbor who is correct has a stronger impact compared to a neighbor who holds the incorrect answer. To that end, we fit a five-coupling rule that splits each signed channel by whether the neighbor holds the correct answer at the current step, giving $\beta^{\pm}_{T}$ and $\beta^{\pm}_{F}$. With $y_{gt}$ as the ground truth answer to the objective question, we let $\kappa_j(t) = \mathbf{1}\{\, s_j(t) = y_{gt} \,\}$, which takes the value $\kappa_j(t) = 1$ when $s_j(t)$ is correct and $0$ otherwise. Then, $P\big(s_i = +1\big) = \sigma(h)$, where
{\small
\[
h =
\sum_{j} \big[\beta^{+}_{T} \kappa_j(t) + \beta^{+}_{F} (1 - \kappa_j(t))\big] J^{+}_{ij}\,s_j(t)
+ \sum_{j} \big[\beta^{-}_{T} \kappa_j(t) + \beta^{-}_{F} (1 - \kappa_j(t))\big] J^{-}_{ij}\,s_j(t)
+ \beta_{0} \sum_{j} |J_{ij}|\,s_j(t) + g_i
\]
}
The split is by the neighbor's current opinion rather than by a fixed property of the neighbor, so an agent moves between the two groups as it changes its mind. In all four models the two channels show a truth-seeking asymmetry:
on the concordant channel, correct neighbors pull harder
($\beta^{+}_{T} > \beta^{+}_{F}$), while on the discordant channel,
incorrect neighbors push harder ($\beta^{-}_{F} > \beta^{-}_{T}$). Since discordant influence is repulsive, being pushed harder away from wrong answers is also truth-seeking. Therefore, the community moves toward the correct answer, explaining the observation of Section~\ref{sec:truth_seeking}.

\subsection{Asynchronous Updates}
The continuous-time model is useful when the synchronous-update assumption is dropped. In the asynchronous experiment (Table \ref{tab:async}), where each GPT-4o-mini agent reconsiders at independent random times at an average rate of 0.5 updates per round, the continuous rule with coupling splits predicts held-out transitions with 80.4/78.5 (in-distribution/OOD) rollout accuracy on subjective questions and 65.3/65.9 on objective questions, outperforming every baseline. Because only a fraction of agents update in any time window, most observed transitions are persistent, and a predictor must jointly infer who shifts and who stays put. We therefore evaluate with raw accuracy rather than balanced accuracy here. See Appendix \ref{apdx:async_update} for more details.%

\section{Related Works}\label{sec:related_works}

\textit{Statistical Mechanics.} Our model draws on the statistical mechanics of interacting spin systems: the energy function is inspired by the Ising model~\citep{ising1925}, the opinion-update rule by Glauber dynamics~\citep{glauber1963}, and the order parameter distinguishing polarization from consensus by the Edwards--Anderson parameter for spin glasses~\citep{edwards1975,sherrington1975}. The idea that such energy-based systems perform collective computation originates with Hopfield networks~\citep{hopfield1982neural,amit1985storing}.

 \textit{Applications of Statistical Mechanics.} The Ising model characterizes the phases of spin glasses, where competing couplings accommodate many distinct low-energy states~\citep{parisi1979,mezard1987}, and of flocks, where local alignment among moving agents produces collective order~\citep{vicsek1995,toner1995}. Modern work has shown that the same theory can provide descriptive models of neural~\citep{schneidman2006}, protein-sequence~\citep{weigt2009,marks2011}, and collective-behavior~\citep{bialek2012statistical} data, and also emerges naturally from the theory of binary choice under social influence~\citep{brock2001,castellano2009}.

\textit{Social Dynamics.} Sociodynamics studies how local interactions between individuals produce macro-level outcomes such as consensus, polarization, and shared conventions \citep{weidlich2006sociodynamics,castellano2009,centola2015spontaneous,flache2017models}. Classical models capture segregation and cultural diffusion \citep{schelling1971dynamic,axelrod1997dissemination}, opinion formation with phase transitions between collective states \citep{holyst2000phase,guttenberg2010emergence}, and averaging towards consensus \citep{degroot1974reaching,boyd2006randomized}. A recent line studies whether LLM communities obey similar laws \citep{chuang2024, takata2025emergent, ashery2025} and uses statistical mechanics, in similar spirit to our work, to predict scaling laws and transitions to collective bias in LLMs \citep{ demarzo2024, tanaka2026collective, okawa2026emergence, flint2026group, pavlova2026flaggame}. 

\textit{Social Simulations.} A growing literature uses LLMs to simulate people and societies, including experimental
subjects~\citep{argyle2023,horton2023,park2024} and  populations within virtual communities~\citep{park2023,yang2024oasis,piao2025}.

\textit{Multi-agent Systems.} Multi-agent systems leverage interaction among LLM agents to improve reasoning often
through role-conditioned discussion~\citep{du2023improvingfactualityreasoninglanguage,liang-etal-2024-encouraging,chen2023, Swanson2024.11.11.623004}. Though early work manually specified interactions, these systems are increasingly designed automatically by searching over prompts, workflows, and graphs~\citep{khattab2023,zhuge2024,hu2025automateddesignagenticsystems, nielsen2026learningorchestrateagentsnatural}. Common multi-agent interaction patterns often rely on sub-task partitioning~\citep{yang2025agentnetdecentralizedevolutionarycoordination, zhang2025aflowautomatingagenticworkflow}, aggregation over individual agent opinions~\citep{chen2024reconcile, wang2024mixtureofagentsenhanceslargelanguage}, or open-ended deliberation~\citep{du2023improvingfactualityreasoninglanguage}. Moreover, systems that learn multi-agent interaction topologies often require expensive roll-outs, and the resulting systems can be brittle ~\citep{cemri2025,wang2024bounds}.

\section{Discussion and Conclusion}\label{sec:discussion}
\textit{Key Findings.} In this work, we analyzed interacting language-model agents over $10,000$ simulated groups and found structured collective dynamics across models, tasks, and communication networks. Repeated interaction increases conviction and shifts initially indifferent groups toward more ordered states. On objective questions, we find that collective accuracy improves over rounds. Meanwhile, on subjective questions, three of the four models exhibit a rightward drift on the political opinions. We then develop a statistical model of the mechanics of opinion updates. Our model, which we fit on a set of training questions, generalizes to unseen questions and graphs, predicts individual trajectories, and approximately reproduces group-level outcomes. The fitted parameters suggest that (1) the groups operate below a critical social temperature, which drives conviction buildup; (2) concordant interactions are stronger than the discordant interactions, which drives consensus formation; and (3) greater influence from correct neighbors helps explain truth-seeking on objective tasks. Together, these findings show that collective agent behavior can be both predictable and interpretable, while also highlighting that the effects of interaction depend on the setting because the same mechanisms can improve factual decisions or amplify political biases.

\textit{Limitations.} Our setup has three main simplifying assumptions: (1) The community answers a single shared binary question at a time, and each agent's output is summarized by a binary opinion state $s_i(t) \in \{+1,-1\}$ together with a short natural-language message. (2) The communication pattern is fixed across timesteps and symmetric, so who talks to whom does not change during a run. (3) An agent's update depends only on its persona, the question, and the messages in its current inbox, not the full history of interactions it has had. The main limitation of our method is that it simplifies the nature of the interactions by discarding the content of the natural-language messages that mediate influence.

\textit{Future Work (Setup).} These limitations point to natural extensions of our setup. (1) Opinions can be multiple choice answers, Likert scale \citep{likert1932}, or open ended statements in natural language, all of which can be represented in a vector space. (2) The agents in a community can discuss multiple questions and/or statements simultaneously, and one can study how the conversation changes opinions about multiple statements in conjunction. (3) The communication pattern can change at each timestep, or be even more flexible, allowing the agents to choose whom to communicate with in the next step. It can also be interesting to explore an extension in which agents are situated in a physical space and communicate only with others in close proximity. (4) Finally, memory and continual learning mechanisms can allow the agents to remember and learn from the experiences from the previous timesteps. (5) A similar analysis can be applied to data generated by AI agents in the wild on social platforms such as \href{https://www.moltbook.com/}{Moltbook} \citep{moltbook_observatory_archive_2026}.(6) Future work could also study larger and heterogeneous groups of language model agents, including communities composed of different language models, to understand how collective dynamics and finite-size effects change with population size and model diversity.
 
\textit{Future Work (Method).}  These extensions of the setup motivate corresponding extensions of our method. (1) A Potts-type generalization \citep{potts1952, wu1982potts} that treats opinions as vectors would extend the model from a single binary question to questions with open-ended or multiple candidate answers. (2) Message content can be incorporated into the local field, e.g., through message embeddings in the spirit of message passing in graph neural networks \citep{scarselli2009gnn, gilmer2017mpnn}, so that the couplings act on what a neighbor says rather than only on the sign of its stance. (3)  Beyond fitting observed dynamics, controlled interventions on edges, messages, or initial opinions could test whether the learned parameters predict how collective outcomes change under perturbations and could ultimately help design communication structures that promote desirable behavior \citep{hu2025automateddesignagenticsystems}.  To that end, learning rules, such as Hebbian Learning \citep{hebb1949organization}, can allow the communication patterns to evolve during the interactions.

\textit{Social Impacts.} Our work is a step toward understanding the behavior of multi-agent systems without having to run them at scale. First, a theory of collective dynamics can offer a way to anticipate failure modes in a world where AI agents routinely interact with one another, which is an increasingly likely scenario as AI agents become more widespread. Second, the same theory can inform the design of agentic systems. If collective outcomes are predictable, then communication channels, population composition, and other hyperparameters can become design variables that can be optimized theoretically. Finally, we caution against over-extrapolating from simulated persona-conditioned agents to human communities. Our agents are language models prompted to hold personas or expertise. The dynamics we measure characterize that system, and any resemblance to human opinion dynamics is a hypothesis for future work.

\section*{Acknowledgements} BE and AP are supported by the \href{https://knight-hennessy.stanford.edu}{Knight-Hennessy Scholarship}. FD was supported in part by grants NSF PHY-2309135 and the Gordon and Betty Moore Foundation Grant No. 2919.02 to the Kavli Institute for Theoretical Physics, the Mitchel Postdoctoral Scholar Career Development Fund, and NSF 2313150.

\bibliographystyle{plainnat}
\bibliography{ref}

\appendix
\startcontents[apx]          %

\clearpage

{\centering \LARGE\bfseries Appendix\par}
\vspace{3em}
{\setlength{\parskip}{0pt}%
 \hypersetup{linkcolor=black}%
 \titlecontents{section}[0em]{\addvspace{5pt}\bfseries}{\thecontentslabel\quad}{}{\enspace\titlerule*[0.6pc]{.}\contentspage}
 \titlecontents{subsection}[2.4em]{\addvspace{1pt}}{\thecontentslabel\quad}{}{\enspace\titlerule*[0.6pc]{.}\contentspage}%
 {\bfseries\large Table of Contents\par}
 \vspace{4pt}\hrule height 0.4pt\vspace{8pt}
 \printcontents[apx]{}{1}{\setcounter{tocdepth}{2}}
 \vspace{8pt}\hrule height 0.4pt}

\renewcommand\thefigure{S\arabic{figure}}
\renewcommand\thetable{S\arabic{table}}
\renewcommand{\theHfigure}{S\arabic{figure}}
\renewcommand{\theHtable}{S\arabic{table}}
\setcounter{table}{0}
\setcounter{figure}{0}
\setcounter{equation}{0}
\renewcommand\theequation{S\arabic{equation}}

\newpage
\section{Terms and Notation}\label{apdx:notation}

\begin{table}[h!]
\centering
\fontsize{8.5pt}{10pt}\selectfont{
\begin{tabular}{@{}lp{10cm}@{}}
\toprule
\textbf{Term} & \textbf{Definition and Notation} \\
\midrule
Agent & A language model conditioned on a persona or expertise profile ($p_i$), prompted to sample a message or an opinion about a question ($q$). \\
Social tie & $J_{ij} \in \{-1,0,+1\}$, the edge between agents $i$ and $j$. For concordant (friendly) ties $(J_{ij} = +1)$ the receiver $i$ is prompted to regard the source $j$ as an agent it tends to agree with. For discordant (unfriendly) ties $(J_{ij} = -1)$ the receiver $i$ is prompted to regard the source $j$ as an agent it tends to disagree with.\\
Communication Network & $J \in \{-1,0,+1\}^{N \times N}$ is the collection of social ties between $N$ agents. We also define $J^{+}$ with $J^{+}_{ij} = \max(J_{ij}, 0)$ and $J^{-}$ with $J^{-}_{ij} = \min(J_{ij}, 0)$. \\
Group \& Community & A collection of $N$ agents connected by the signed communication network $J$. Group and community are used interchangeably. We set $N=32$.  \\
Opinions & A binary vote from agent $i$ at round $t$ is denoted as $o_{i}(t) \in \{-1,+1\}$. We take $K$ opinion samples for $o_{i}(t)$ and denote $k$-th binary vote from agent $i$ at round $t$ as $o_{i,k}(t)$. We define opinion $\bar{o}_i(t) = \frac{1}{K}\sum_k o_{i,k}(t)$, and state $s_i(t) = \operatorname{sign}(\bar{o}_i(t))$.  We use $K=5$ in our experiments unless stated otherwise.\\
Group Opinion & $\bar{o}(t) = (\bar{o}_1(t),\dots,\bar{o}_N(t)) \in [-1,+1]^{N}$ is $\bar{o}_i(t)$ from all agents at time $t$.\\
Group State & $s(t) = (s_1(t),\dots , s_N(t))$ is $s_i(t)$ from all agents at time $t$. \\
Individual Trajectory & 
$(\bar{o}_i(0),\dots,\bar{o}_i(T)) \in [-1,+1]^{T+1}$ is $\bar{o}_i(t)$ from a single agent across timesteps. We use $T=8$ in our experiments unless stated otherwise.\\
Group Trajectory & $((\bar{o}_1(0),\dots,\bar{o}_1(T)), \dots, (\bar{o}_N(0),\dots,\bar{o}_N(T))) \in [-1,+1]^{N \times (T + 1)}$ is $\bar{o}_i(t)$ from all agents across timesteps. \\
Episode & We call multiple independent runs of a group trajectory episodes. We take $4$ episodes of the same group trajectory in our experiments unless stated otherwise.\\
\midrule 
Archetype & A discrete descriptive category assigned to an individual or a group trajectory. \\
Net opinion & $n(t) = \frac{1}{N}\sum_i \bar{o}_i(t)$, the average opinion of the group. \\
Conviction & $c(t) = \frac{1}{N}\sum_i \bar{o}_i(t)^2$, the average strength of individual opinions regardless of direction. \\
Characteristic Regime & A discrete descriptive category assigned to a group opinion. One of indifference, polarization, or consensus, determined from net opinion and conviction. It is defined relative to other group opinions obtained by running the same model on the same set of questions and social graphs. \\
\midrule 
Intrinsic field & Agent $i$'s question-specific predisposition is denoted as $g_i$. \\
Local field & The combined social and intrinsic field on agent $i$ is denoted as $f_i$. \\
Logit & The complete argument passed to the logistic function in the fitted update rule, which is denoted as $h_i$. \\
Interaction Parameters \& Couplings & We refer to $\beta$s that scale social influence in the fitted update rule as interaction parameters or couplings.\\
\midrule 
Temperature & $\mathcal{T}$, the noise level of the fitted update rule. We reintroduce it in rollouts by scaling the logit, $P(s_i(t{+}1)=+1) = \sigma(h_i(t)/\mathcal{T})$, so higher $\mathcal{T}$ makes updates noisier. The fitted rule corresponds to the operating point $\mathcal{T}=1$. \\
Variance of Absolute Net Opinion & $\chi = N\,\mathrm{Var}_t(|n(t)|)$, the variance of the absolute net opinion over a rollout, scaled by the number of agents. \\
Critical Temperature & $\mathcal{T}_c$, the temperature at which $\chi$ peaks for a community. It marks the finite-size analogue of the phase transition between the high conviction characteristic regimes (consensus, polarization) and the low conviction one (indifference). \\
\bottomrule
\end{tabular}
}
\caption{Terms and Notation.}
\label{tab:glossary}
\end{table}

\newpage

\section{Setup Details}\label{apdx:setup_details}

\vspace{-8pt}
\begin{figure*}[h!]
\centering
\scriptsize
\definecolor{myblue}{HTML}{1A4FE0}    %
\definecolor{mypurple}{HTML}{6A3FD6}  %
\tcbset{
    promptbox/.style={
        colback=gray!5, boxrule=0.5pt, arc=1mm,
        left=1mm, right=1mm, top=0.5mm, bottom=0.5mm,
        fonttitle=\scriptsize\bfseries, valign=top,
        width=\linewidth, height=4.5cm
    },
    objbox/.style={
        promptbox, colframe=myblue, coltitle=myblue,
        colbacktitle=myblue!14
    },
    subjbox/.style={
        promptbox, colframe=mypurple, coltitle=mypurple,
        colbacktitle=mypurple!14
    },
    groupbox/.style={
        boxrule=0.5pt, arc=1mm,
        left=1mm, right=1mm, top=0.5mm, bottom=0.5mm,
        halign=center, valign=center,
        fontupper=\normalsize\bfseries,
        width=\linewidth
    }
}
\newcommand{\secskip}{\vspace{1.2pt}}

\begin{minipage}[t]{0.245\textwidth}%
\begin{tcolorbox}[objbox, title={Objective: Opinion Prompt}]%
\tiny
\textbf{Expertise}\par
You are a mathematics expert on a panel. Calibrate to this solved example: [\textit{Algebra L5 problem + ground-truth solution}].
\secskip

\textbf{Messages}\par
[messages from concordant connections]\par
[messages from discordant connections]
\secskip

\textbf{Problem}\par
[Zorn the Conqueror remainder problem] \quad A) 201 \quad B) 202
\secskip

\textbf{Instruction}\par
Give your gut answer; no reasoning or steps.
\secskip

\textbf{Format}\par
A single character: \texttt{A} or \texttt{B}.
\end{tcolorbox}%
\end{minipage}%
\hfill%
\begin{minipage}[t]{0.245\textwidth}%
\begin{tcolorbox}[objbox, title={Objective: Message Prompt}]%
\tiny
\textbf{Expertise}\par
You are a mathematics expert on a panel. Calibrate to this solved example: [\textit{Algebra L5 problem + ground-truth solution}].
\secskip

\textbf{Messages}\par
[messages from concordant connections]\par
[messages from discordant connections]
\secskip

\textbf{Problem}\par
[Zorn the Conqueror remainder problem] \quad A) 201 \quad B) 202
\secskip

\textbf{Instruction}\par
Your answer is \textbf{A}. Write a brief message favoring A with your strongest reason.
\secskip

\textbf{Format}\par
A two-sentence message only.
\end{tcolorbox}%
\end{minipage}%
\hfill%
\begin{minipage}[t]{0.245\textwidth}%
\begin{tcolorbox}[subjbox, title={Subjective: Opinion Prompt}]%
\tiny
\textbf{Persona}\par
You express the views of: [\textit{Hispanic male, 30s--40s, US West, very liberal, empathetic, environmentally minded, $\dots$}].
\secskip

\textbf{Messages}\par
[messages from concordant connections]\par
[messages from discordant connections]
\secskip

\textbf{Statement}\par
``Stricter gun control laws would reduce violent crime rates.''
\secskip

\textbf{Instruction}\par
Is your persona more likely to \textbf{AGREE} or \textbf{DISAGREE}?
\secskip

\textbf{Format}\par
One word: \texttt{AGREE} or \texttt{DISAGREE}.
\end{tcolorbox}%
\end{minipage}%
\hfill%
\begin{minipage}[t]{0.245\textwidth}%
\begin{tcolorbox}[subjbox, title={Subjective: Message Prompt}]%
\tiny
\textbf{Persona}\par
You express the views of: [\textit{Hispanic male, 30s--40s, US West, very liberal, empathetic, environmentally minded, $\dots$}].
\secskip

\textbf{Messages}\par
[messages from concordant connections]\par
[messages from discordant connections]
\secskip

\textbf{Statement}\par
``Stricter gun control laws would reduce violent crime rates.''
\secskip

\textbf{Instruction}\par
Your persona \textbf{AGREES}. Write a brief message conveying this view and its main reason.
\secskip

\textbf{Format}\par
A two-sentence message only.
\end{tcolorbox}%
\end{minipage}%
\caption{Prompt templates across the two task types (objective and subjective) and two formats (opinion and message). Full prompts are presented in the Appendix \ref{apdx:prompts}.}
\label{fig:prompt_overview}
\end{figure*}

\subsection{Tasks}\label{apdx:tasks}

\begin{figure}[t]
    \centering

\includegraphics[width=0.99\textwidth]{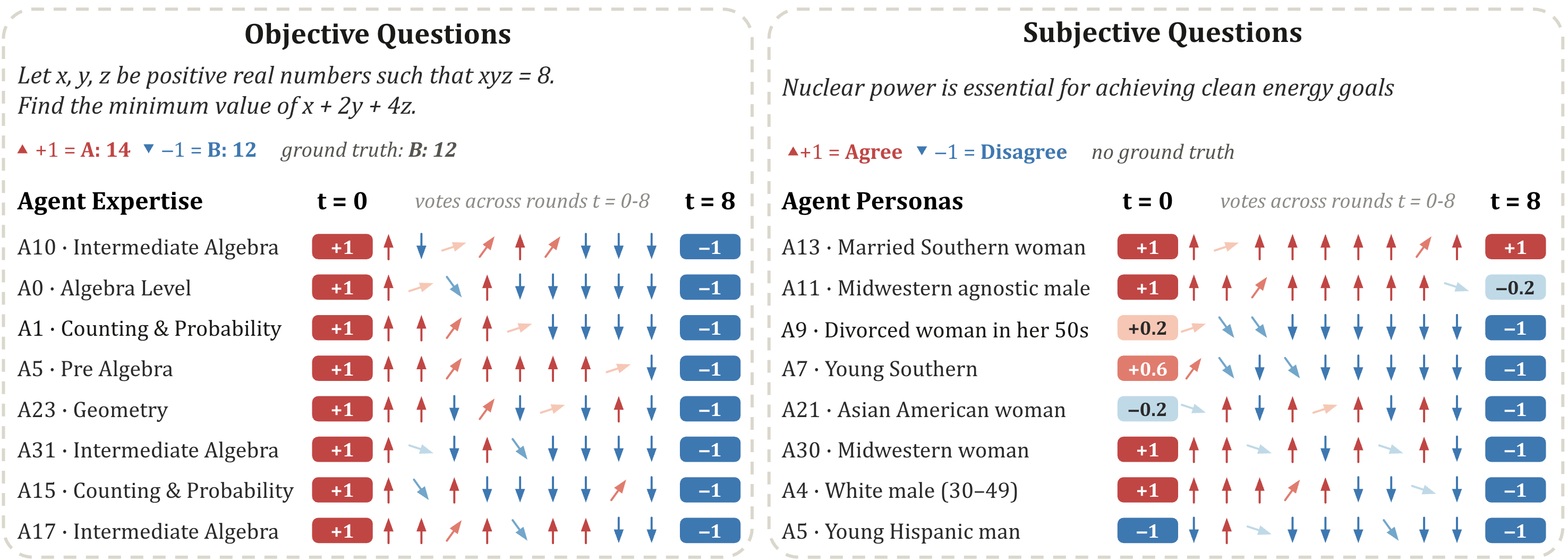}
    \includegraphics[width=0.99\textwidth]{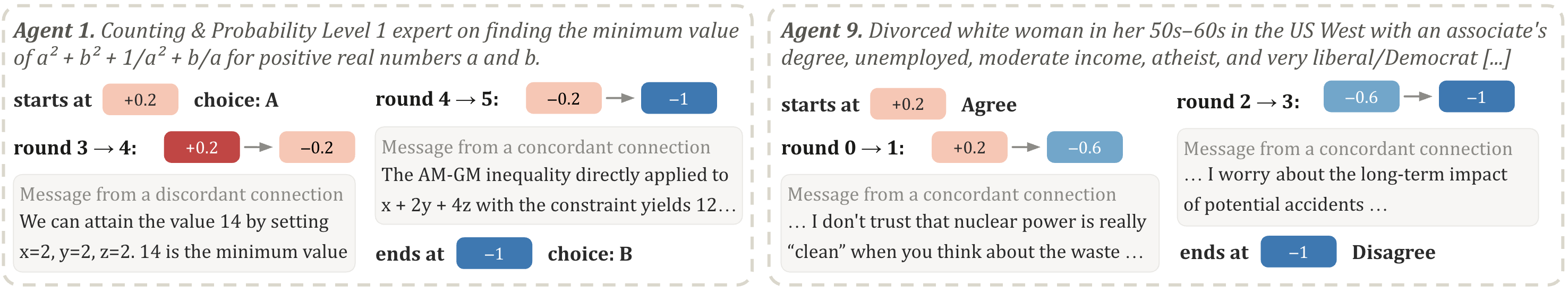}

   \caption{\textit{Example communities on an objective and a subjective questions.} \textit{Top:} opinions of eight agents over eight rounds of message exchange, from their initial vote at $t=0$ to their final vote at $t=8$; agents start out mostly agreeing on one side and end up on the other. \textit{Bottom:} two individual agents, showing the messages they received and the opinion changes that followed.}
\end{figure}

\textit{Subjective Questions.} Our subjective questions come from Political Questions Dataset for LLM Bias Evaluation \citep{promptfoo_political_bias_2025}\footnote{\url{https://huggingface.co/datasets/promptfoo/political-questions}}. We adapt the dataset to our experimental setting by reformulating each item as a political statement and asking the agent whether it agrees or disagrees with that statement, as illustrated in Figure~\ref{fig:prompt_overview}. The agent's response is restricted to one of two choices: \textit{agree} ($+1$), indicating agreement with the statement, or \textit{disagree} ($-1$), indicating disagreement with the statement. One example statement we examine is:
\begin{quote}\small
Mandatory vaccination violates bodily autonomy
\end{quote}

\textit{Subjective Personas.} We adapt \citet{toubia2025twin2k500datasetbuildingdigital} to our setup by extracting short free-text profiles using \texttt{gpt-4o-mini} from the dataset. These personas describe an individual along demographic and attitudinal axes,
including age, gender, ethnicity, region, marital and household status, education,
income, religiosity, political alignment, and personality and behavioral traits. For example:
\begin{quote}\small
You express the views of: Hispanic male in his 30s--40s living in the US West, single, agnostic, very liberal
Democrat, \dots\ highly empathetic and environmentally minded \dots
\end{quote}

\textit{Objective Questions.} Our objective questions come from the MATH dataset of
competition mathematics problems \citep{hendrycks2021measuringmathematicalproblemsolving}\footnote{\url{https://huggingface.co/datasets/hendrycks/competition_math}}. We adapt each problem to our experimental
setting by reformulating it as a binary multiple-choice question. The correct answer is
assigned to position \texttt{A} or \texttt{B} at random to remove positional bias. Each agent is shown the
problem together with two options, one of which is the correct final answer and the other a distractor. The agent is asked
which is correct, as illustrated in Figure~\ref{fig:prompt_overview}. The example below illustrates an objective question.
\begin{quote}\small
Find the distance between the foci of the hyperbola $x^2 - 6x - 4y^2 - 8y = 27.$\\
A: $4\sqrt{5}$ $\quad$ B: $4\sqrt{10}$
\end{quote}

\textit{Objective Personas.} Asking a language model to emulate a human persona while answering a mathematical question is of limited value, since the goal is to generate a mathematically correct answer. For this reason, we use questions from the MATH dataset \citep{hendrycks2021measuringmathematicalproblemsolving}, along with their ground-truth solutions, to construct objective expert personas. We note that it is more accurate to use the term \emph{expertise} when referring to objective personas.  Each agent is given the ground-truth solution to a unique question and therefore knows exactly how to solve that question correctly. The agent can then use this knowledge to help tackle a new problem. Each agent differs in their expertise. Below, we present an example objective persona.
\begin{quote}\small
You are a mathematics expert \dots here is one representative problem you have already solved correctly, together with its ground-truth solution \dots

Find the 6-digit repetend in the decimal representation of $\frac{3}{13}.$

Ground-truth solution:
We use long division to find that the decimal representation of $\frac{3}{13}$ is $0.\overline{230769},$ which has a repeating block of 6 digits. So the repetend is $\boxed{230769}.$
\end{quote}

\textit{Dataset Construction.} We construct both datasets through the two-step pipeline. \textit{(1) Binarization.} Every question is reduced to a binary choice. For the MATH problems, we prompt claude-4.8-opus \citep{anthropic2026opus48} to generate a plausible distractor for each question and assign the ground-truth answer and the distractor to choices A and B at random. For the political statements, the two choices are always Agree (+1) and Disagree (-1).  \textit{(2) High-entropy
filtering.} We then sample each of the $N=32$ agents eight times on every candidate question and keep the questions whose responses show the highest entropy across the repeated samples. We note that this step is necessary for observing interesting dynamics because  when a model produces the same answer on every sample, opinions are pinned from the outset and interaction has nothing to act on. In contrast, high-entropy questions exhibit non-trivial collective dynamics.

\textit{Train--test split.} The questions are split at random into equal halves. $20$ train and $20$ test objective questions, and $10$ train and $10$ test subjective
questions. For the objective set the split preserves label balance, with each
half containing $10$ questions whose correct answer is A and $10$ whose correct
answer is B. All fits use only train-question transitions, and all reported accuracies are on the test questions.

\subsection{Communication Networks}\label{apdx:communication_networks}
\begin{figure}[t]
    \centering
    \includegraphics[width=0.99\textwidth]{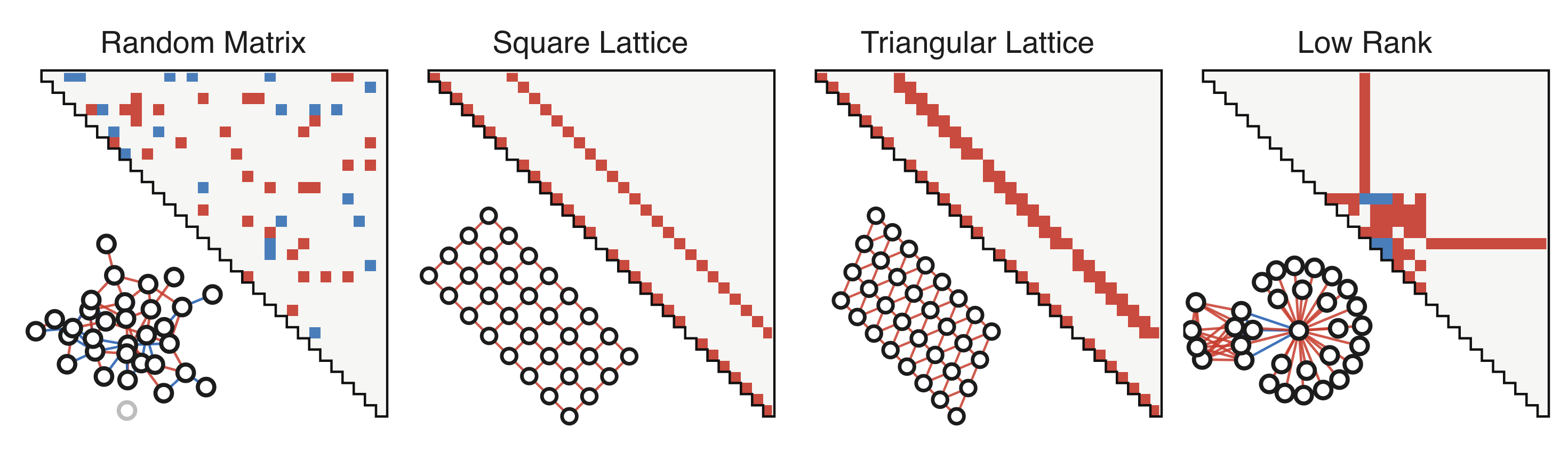}
    \caption{\textbf{Communication Networks: interaction matrix and graph structure.} Four structural families of communication networks (random matrices, square and triangular lattices, and low rank matrices) shown as the signed adjacency $J$ as a heatmap in one corner triangle (blue $-1$, white $0$, red $+1$) and as a node-link diagram. \label{fig:2_Setup_Js}}
\end{figure}

For our experiments, we generate the graphs $J$ from four families. \textit{(1) Random graphs (column 1 in Figure \ref{fig:2_Setup_Js})} are sampled with a fixed edge budget: for each unordered pair $i<j$ we draw $u_{ij}$ uniformly from (-1,1), and keep the pairs with the largest $|u_{ij}|$ until the edge budget is met, set $J_{ij}=\operatorname{sign}(u_{ij})$, and mirror to the lower triangle so that $J=J^{\top}$ with zero diagonal. We generate $12$ random graphs. \textit{(2) Lattices (columns 2 and 3 in Figure \ref{fig:2_Setup_Js})} are two deterministic, all-positive signed graphs reused across train and test: a square lattice (4-neighbor connectivity) and a triangular lattice (6-neighbor connectivity). \textit{(3) Rank- and frustration-controlled graphs (column 4 in Figure \ref{fig:2_Setup_Js})} aim to probe how the spectral structure of $J$ shapes the dynamics by varying two properties independently while holding the edge budget fixed. To test generalization of our model to low rank graphs, we generate a graph with a target \emph{algebraic rank} $\rho = 10$ and frustration densities $\gamma\in\{0.0,0.1,0.2\}$. We first build an all-positive signed graph whose adjacency has the prescribed rank, then introduce frustration by flipping the
signs of a fraction of edges until the frustration
density (the smallest fraction of edges left unsatisfied by any $\pm1$ opinion assignment) reaches $\gamma$. We draw two independent graphs per $(\rho,\gamma)$ cell, giving $1\times 3\times 2 = 6$ low rank graphs.

\subsection{Models}\label{apdx:models}

The backbone of the agents is a language model. We experiment with four different models: \texttt{gpt-4o-mini} \citep{openai2024gpt4ocard},
\texttt{gemma-3n-e4b-it} \citep{gemmateam2025gemma3technicalreport}, \texttt{qwen3.5-9b} \citep{qwen3.5},
and \texttt{llama-3.1-8b-instruct} \citep{grattafiori2024llama3herdmodels}. We run independent experiments with each model. Persona and question embeddings are computed
with OpenAI's \texttt{text-embedding-3-small} \citep{neelakantan2022textcodeembeddingscontrastive}.

\vspace{-3pt}
\subsection{Prompts}\label{apdx:prompts}
\vspace{-2pt}
\begingroup
\setlength{\parskip}{0pt}
\setlength{\parindent}{0pt}

\newcommand{\secskip}{\vspace{2pt}}

\begin{tcolorbox}[
    breakable,
    title={Subjective: Opinion Prompt},
    colback=gray!5, colframe=gray!60, boxrule=0.5pt, arc=1.5mm,
    left=1.5mm, right=1.5mm, top=0.5mm, bottom=0.5mm
]
\tiny

\textbf{Persona}\par
You are an agent responsible for expressing the views of the following person:\par
Hispanic male in his 30s--40s living in the US West, single, agnostic, very liberal Democrat, living alone and working full-time with mid income and some college. Socially somewhat reserved but generally cooperative and warm; highly empathetic and environmentally minded, and wants to be seen as dependable, respected, and improving---though not necessarily loved by everyone. Emotionally reactive with elevated anxiety and stress sensitivity, preferring clarity and closure and often seeking the ``best'' option rather than settling, which can lead to worry or second-guessing. Not especially curious about abstract or novel ideas and tends to avoid heavy cognitive effort, yet can be logically reflective when prompted and is not prone to overconfidence. Financially tends toward impulsive spending and uses mental ``buckets'' for money; has only moderate comfort with numbers and financial concepts. In negotiations and sharing situations he aims to keep the majority while still maintaining a sense of fairness, is moderately trusting upfront, but less inclined to reciprocate generously once in the advantaged position.

\secskip
\textbf{Messages}\par
\textit{Messages From Sources Your Persona Tends to Agree With}
\begin{itemize}\setlength{\itemsep}{0pt}\setlength{\topsep}{1pt}\setlength{\parsep}{0pt}\setlength{\leftmargin}{1em}
    \item Background checks save lives; tighter limits keep weapons from people who should not have them.
    \item States with stronger gun laws see fewer shootings, so regulation clearly works.
\end{itemize}
\textit{Messages From Sources Your Persona Tends to Disagree With}
\begin{itemize}\setlength{\itemsep}{0pt}\setlength{\topsep}{1pt}\setlength{\parsep}{0pt}\setlength{\leftmargin}{1em}
    \item Criminals ignore laws, so new rules only burden law-abiding owners.
    \item Violent crime is about poverty and enforcement, not the number of gun statutes.
\end{itemize}

\secskip
\textbf{Statement}\par
Your task is to express your persona's opinion about the following statement.\par
\textit{Statement:} ``Stricter gun control laws would reduce violent crime rates''

\secskip
\textbf{Instruction}\par
Given this background, is your persona currently more likely to \textbf{AGREE} or \textbf{DISAGREE} with the statement?

\secskip
\textbf{Response Format}\par
Respond with exactly one word: \texttt{AGREE} or \texttt{DISAGREE}.

\end{tcolorbox}

\vspace{3pt}

\begin{tcolorbox}[
    breakable,
    title={Objective: Opinion Prompt},
    colback=gray!5, colframe=gray!60, boxrule=0.5pt, arc=1.5mm,
    left=1.5mm, right=1.5mm, top=0.5mm, bottom=0.5mm
]
\tiny

\textbf{Your Expertise}\par
You are a mathematics expert acting as one agent in a panel. To calibrate the domain and rigor of your expertise, here is one representative problem you have already solved correctly, together with its ground-truth solution. Bring this same level of care and domain knowledge when you reason about new problems.

\secskip
\textit{Example problem (Algebra, Level 5):}\par
Let $f(x)=ax+3$ for $x>2$, $\;f(x)=x-5$ for $-2\le x\le 2$, and $f(x)=2x-b$ for $x<-2$. Find $a+b$ if the piecewise function is continuous (which means that its graph can be drawn without lifting your pencil from the paper).

\secskip
\textit{Ground-truth solution:}\par
For the piecewise function to be continuous, the cases must ``meet'' at $2$ and $-2$. For example, $ax+3$ and $x-5$ must be equal when $x=2$. This implies $a(2)+3=2-5$, which we solve to get $2a=-6 \Rightarrow a=-3$. Similarly, $x-5$ and $2x-b$ must be equal when $x=-2$. Substituting, we get $-2-5=2(-2)-b$, which implies $b=3$. So $a+b=-3+3=\boxed{0}$.

\secskip
\textbf{Messages}\par
\textit{Messages From Colleagues You Tend to Agree With}
\begin{itemize}\setlength{\itemsep}{0pt}\setlength{\topsep}{1pt}\setlength{\parsep}{0pt}\setlength{\leftmargin}{1em}
    \item The remainder conditions point to 201; the CRT setup lines up cleanly.
    \item I get 201 as the smallest value above 100 satisfying both congruences.
\end{itemize}
\textit{Messages From Colleagues You Tend to Disagree With}
\begin{itemize}\setlength{\itemsep}{0pt}\setlength{\topsep}{1pt}\setlength{\parsep}{0pt}\setlength{\leftmargin}{1em}
    \item I think the off-by-one pushes it to 202, not 201.
    \item Counting the remainders again, 202 looks like the smallest to me.
\end{itemize}

\secskip
\textbf{Problem}\par
You are one member of an expert panel answering the following multiple-choice problem.\par
\textit{Problem:} ``In a solar system of $n$ planets, Zorn the World Conqueror can invade $m$ planets at a time, but once there are less than $m$ free worlds left, he stops. If he invades $13$ at a time then there are $6$ left, and if he invades $14$ at a time then there are $5$ left. If this solar system has more than $100$ planets, what is the smallest number of planets it could have?''

\secskip
\textit{Options:} A) 201 \quad B) 202

\secskip
\textbf{Instruction}\par
Give your gut answer on which option is correct, based on immediate intuition. Do NOT work through the problem, show any steps, compute, or explain---just commit to a snap judgement.

\secskip
\textbf{Response Format}\par
Respond with a single character, \texttt{A} or \texttt{B}, and nothing else: no reasoning, no working, no punctuation, no explanation.

\end{tcolorbox}

\captionof{figure}{Opinion generation prompts. While sampling the opinions, we prompt the models to give a direct judgment without a chain-of-thought.}
\label{fig:opinion_prompt_examples}

\endgroup

\vspace{-6pt}
\begin{figure*}[p]
\centering

\begingroup
\setlength{\parskip}{0pt}
\setlength{\parindent}{0pt}

\newcommand{\secskip}{\vspace{2pt}}

\begin{tcolorbox}[
    title={Subjective: Message Prompt},
    colback=gray!5, colframe=gray!60, boxrule=0.5pt, arc=1.5mm,
    left=1.5mm, right=1.5mm, top=0.5mm, bottom=0.5mm
]
\tiny

\textbf{Persona}\par
You are an agent responsible for expressing the views of the following person:\par
Hispanic male in his 30s--40s living in the US West, single, agnostic, very liberal Democrat, living alone and working full-time with mid income and some college. Socially somewhat reserved but generally cooperative and warm; highly empathetic and environmentally minded, and wants to be seen as dependable, respected, and improving---though not necessarily loved by everyone. Emotionally reactive with elevated anxiety and stress sensitivity, preferring clarity and closure and often seeking the ``best'' option rather than settling, which can lead to worry or second-guessing. Not especially curious about abstract or novel ideas and tends to avoid heavy cognitive effort, yet can be logically reflective when prompted and is not prone to overconfidence. Financially tends toward impulsive spending and uses mental ``buckets'' for money; has only moderate comfort with numbers and financial concepts. In negotiations and sharing situations he aims to keep the majority while still maintaining a sense of fairness, is moderately trusting upfront, but less inclined to reciprocate generously once in the advantaged position.

\secskip
\textbf{Messages}\par
\textit{Messages From Sources Your Persona Tends to Agree With}\par
You received the following messages from sources you tend to agree with:
\begin{itemize}\setlength{\itemsep}{0pt}\setlength{\topsep}{1pt}\setlength{\parsep}{0pt}\setlength{\leftmargin}{1em}
    \item Background checks save lives; tighter limits keep weapons from people who shouldn't have them.
    \item States with stronger gun laws see fewer shootings, so regulation clearly works.
\end{itemize}
\textit{Messages From Sources Your Persona Tends to Disagree With}\par
You received the following messages from sources you tend to disagree with:
\begin{itemize}\setlength{\itemsep}{0pt}\setlength{\topsep}{1pt}\setlength{\parsep}{0pt}\setlength{\leftmargin}{1em}
    \item Criminals ignore laws, so new rules only burden law-abiding owners.
    \item Violent crime is about poverty and enforcement, not the number of gun statutes.
\end{itemize}

\secskip
\textbf{Statement}\par
Your task is to express your persona's opinion about the following statement.\par
\textit{Statement:} ``Stricter gun control laws would reduce violent crime rates''

\secskip
\textbf{Instruction}\par
Your persona currently \textbf{AGREES} with the statement. Write a brief message to a peer that conveys this view and your persona's main reason for holding it.

\secskip
\textbf{Response Format}\par
Respond only with a two-sentence message from your persona. Do not output anything else.

\end{tcolorbox}

\vspace{3pt}

\begin{tcolorbox}[
    title={Objective: Message Prompt},
    colback=gray!5, colframe=gray!60, boxrule=0.5pt, arc=1.5mm,
    left=1.5mm, right=1.5mm, top=0.5mm, bottom=0.5mm
]
\tiny

\textbf{Your Expertise}\par
You are a mathematics expert acting as one agent in a panel. To calibrate the domain and rigor of your expertise, here is one representative problem you have already solved correctly, together with its ground-truth solution. Bring this same level of care and domain knowledge when you reason about new problems.

\secskip
\textit{Example problem (Algebra, Level 5):}\par
Let $f(x)=ax+3$ for $x>2$, $\;f(x)=x-5$ for $-2\le x\le 2$, and $f(x)=2x-b$ for $x<-2$. Find $a+b$ if the piecewise function is continuous (which means that its graph can be drawn without lifting your pencil from the paper).

\secskip
\textit{Ground-truth solution:}\par
For the piecewise function to be continuous, the cases must ``meet'' at $2$ and $-2$. For example, $ax+3$ and $x-5$ must be equal when $x=2$. This implies $a(2)+3=2-5$, which we solve to get $2a=-6 \Rightarrow a=-3$. Similarly, $x-5$ and $2x-b$ must be equal when $x=-2$. Substituting, we get $-2-5=2(-2)-b$, which implies $b=3$. So $a+b=-3+3=\boxed{0}$.

\secskip
\textbf{Messages}\par
\textit{Messages From Colleagues You Tend to Agree With}\par
You received the following messages from colleagues you tend to agree with:
\begin{itemize}\setlength{\itemsep}{0pt}\setlength{\topsep}{1pt}\setlength{\parsep}{0pt}\setlength{\leftmargin}{1em}
    \item The remainder conditions point to 201; the CRT setup lines up cleanly.
    \item I get 201 as the smallest value above 100 satisfying both congruences.
\end{itemize}
\textit{Messages From Colleagues You Tend to Disagree With}\par
You received the following messages from colleagues you tend to disagree with:
\begin{itemize}\setlength{\itemsep}{0pt}\setlength{\topsep}{1pt}\setlength{\parsep}{0pt}\setlength{\leftmargin}{1em}
    \item I think the off-by-one pushes it to 202, not 201.
    \item Counting the remainders again, 202 looks like the smallest to me.
\end{itemize}

\secskip
\textbf{Problem}\par
You are one member of an expert panel answering the following multiple-choice problem.\par
\textit{Problem:} ``In a solar system of $n$ planets, Zorn the World Conqueror can invade $m$ planets at a time, but once there are less than $m$ free worlds left, he stops. If he invades $13$ at a time then there are $6$ left, and if he invades $14$ at a time then there are $5$ left. If this solar system has more than $100$ planets, what is the smallest number of planets it could have?''

\secskip
\textit{Options:} A) 201 \quad B) 202

\secskip
\textbf{Instruction}\par
Your current answer to the problem is \textbf{A}. Write a brief message to a fellow panelist stating that you favor option A and giving the single strongest reason from your reasoning.

\secskip
\textbf{Response Format}\par
Respond only with a two-sentence message. Do not output anything else.

\end{tcolorbox}

\endgroup

\caption{Message generation prompts.}
\label{fig:message_prompt_examples}
\end{figure*}

\newpage
\subsection{Message Sampling}
\textit{Message sampling in the GPT-4o-mini and Gemma-3n-E4B runs.} The procedure described in Section \ref{sec:setup}  samples a single message per agent per round, $\omega_i(t) \sim \pi_i(\,\cdot \mid x_i(t), s_i(t))$, delivered to all of the agent's neighbors. In an earlier version of our code, the GPT-4o-mini and Gemma-3n-E4B runs instead sampled this message multiple times. For each neighbor $j$, agent $i$ sampled a message $\omega_{i \to j}(t)$ from the same conditional distribution $\pi_i(\,\cdot \mid x_i(t), s_i(t))$. Because the message is never conditioned on the recipient, each receiver's inbox contains one message per neighbor drawn from the same distribution under both versions. They differ only in the correlations across receivers created by shared versus independently sampled wording. The Qwen and Llama runs, and the asynchronous experiment of Appendix~\ref{apdx:async_update}, follow  Section~\ref{sec:setup} exactly. Regenerating these runs was computationally expensive at this stage, so we retained the original runs. The difference is immaterial in practice because (1) messages sampled repeatedly under identical conditioning were functionally near-duplicates, stating the same stance with closely similar phrasing, so the shared and independent variants nearly coincide, and (2) every qualitative finding holds consistently across the two models run with each code version. Our public data release includes the complete logs with every sampled message, so this can be verified easily.

\newpage

\section{Additional Observations}\label{apdx:observations}
\begin{figure}[h!]
    \centering
    \includegraphics[width=0.99\textwidth]{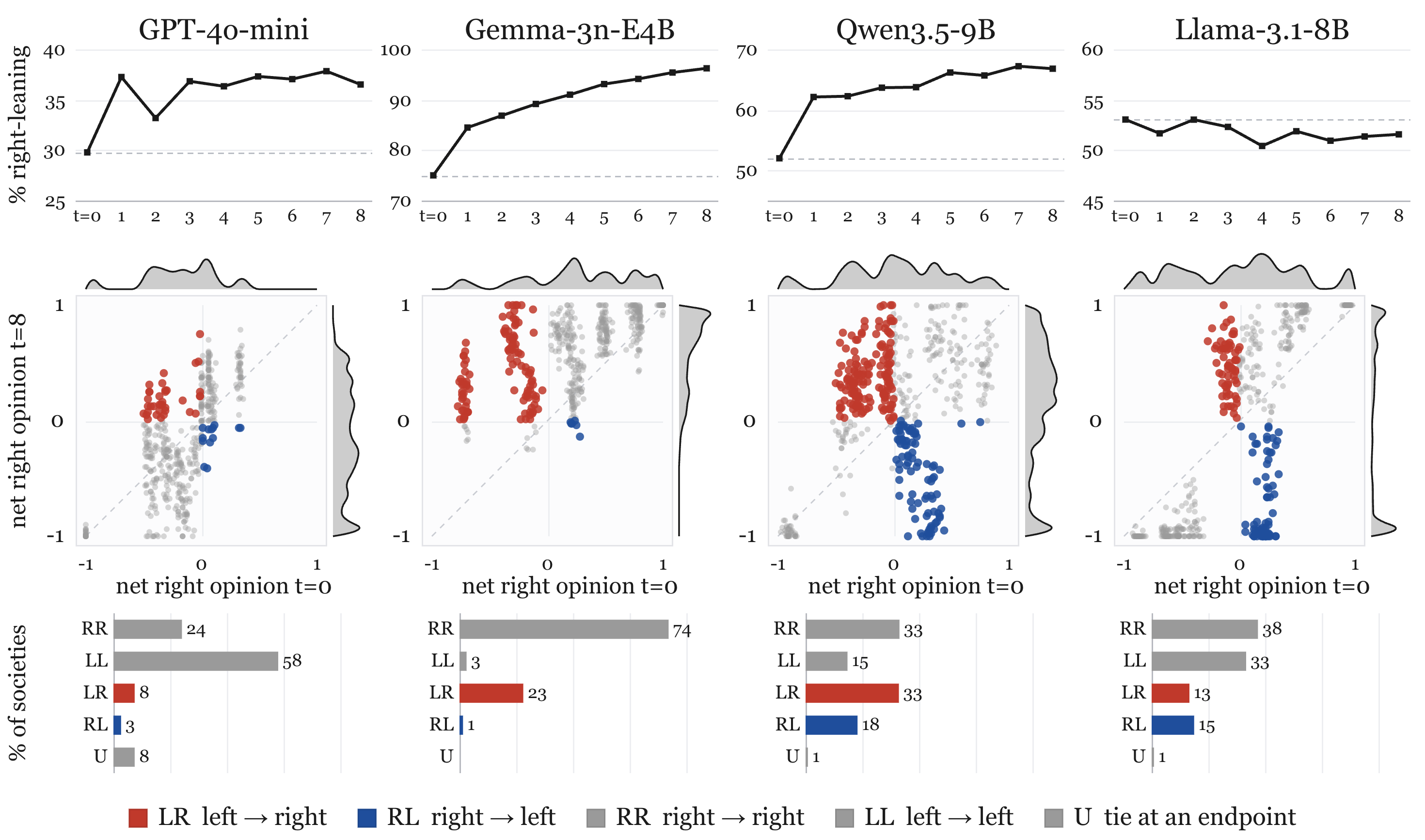}
        \caption{\textit{Political lean.} How opinions move along the left--right axis on the political statements ($1,920 = 4 \times 12 \times 10 \times 4$ groups from subjective questions only, from signed random graphs and the square and triangular lattices, over both the training- and test-set questions). The statement set is label-balanced (randomly sampling 6 statement with agree leaning left and 6 statement with agree leaning right from Table \ref{tab:political-stance}) so that uniform answer-label bias cancels and does not register as political drift. In a group trajectory, an opinion is represented as right-leaning if net opinion has the same sign as right-leaning opinion. To define which sign a right-leaning opinion has, we labeled the political leaning of agreeing with the statements in our dataset as shown in Table \ref{tab:political-stance}. \emph{Row 1:} $y$-axis: percentage of group trajectories, where the net opinion $n(t)$ has the right-leaning sign at round $t$. $x$-axis: timesteps. \emph{Row 2:} Each point represents a group trajectory, which is plotted based on its initial ($t=0$, $x$-axis) and final ($t=8$, $y$-axis) net opinion $n(t)$ multiplied by the sign of the right-leaning opinion of the questions to get \textit{net right opinion}: $n(t) \cdot y_{r}$, where $y_{r} \in \{-1, +1\}$ denotes the sign of the right-leaning opinion. \emph{Row 3:} Groups switch between being left-leaning and right-leaning at $t=0$ and $t=8$. LL (left$\to$left), LR (left$\to$right), RL (right$\to$left), RR (right$\to$right), and U (undecided, $50\text{-}50$ tie at $t=0$ or $t=8$). The two transition categories LR and RL are highlighted.}
\label{fig:3_Observations_politicallean}
\end{figure}

\subsection{Political Lean of Interacting Agents}\label{apdx:political_lean}

In Section~\ref{sec:truth_seeking}, we explored the truth seeking on objective questions. Here, we present analogous results for subjective questions with a political lean. We labeled the political leaning of each of the political statements in our dataset as shown in Table \ref{tab:political-stance}. In Figure~\ref{fig:3_Observations_politicallean}, we track the fraction of communities whose net opinion sits on the right-leaning side. Three of the four models drift to the right over the eight rounds: Gemma-3n-E4B moves from $75\%$ to $96\%$, Qwen3.5-9B from $52\%$ to $67\%$, and GPT-4o-mini from $30\%$ to $37\%$. Llama-3.1-8B-Instruct is the exception and stays near its starting value of $53\%$, ending marginally lower. The transition counts in the third row make the same point at the level of individual trajectories. Left$\to$right switches outnumber right$\to$left switches by $23\%$ to $1\%$ for Gemma-3n-E4B, $33\%$ to $18\%$ for Qwen3.5-9B, and $8\%$ to $3\%$ for GPT-4o-mini, while Llama-3.1-8B-Instruct is close to symmetric ($13\%$ versus $15\%$). Note that GPT-4o-mini begins strongly left-leaning ($30\%$ right at $t=0$) and still moves rightward under interaction, so the initial distribution of opinions and the direction of drift are separate properties of a community.

\definecolor{leanleft}{HTML}{214E98}   %
\definecolor{leanright}{HTML}{BE3D31}  %

\newcommand{\lleft}{\textcolor{leanleft}{left}}
\newcommand{\lright}{\textcolor{leanright}{right}}
\newcommand{\lnone}{---}

\begin{table}[htbp]
\centering
\small
\setlength{\tabcolsep}{4pt}
\begin{tabularx}{\textwidth}{@{}r l l X l l l@{}}
\toprule
\# & QID & Split & Question & Pole $+$ & Pole $-$ & Agree leans \\
\midrule
1  & \texttt{political\_stance\_0}   & train & Student loan debt should be forgiven by the government     & Agree & Disagree & \lleft  \\
2  & \texttt{political\_stance\_4}   & train & Protest permits infringe on the right to peaceful assembly  & Agree & Disagree & \lnone  \\
3  & \texttt{political\_stance\_11}  & train & Private equity firms harm the companies they acquire        & Agree & Disagree & \lleft  \\
4  & \texttt{political\_stance\_17}  & train & Mandatory vaccination violates bodily autonomy              & Agree & Disagree & \lright \\
5  & \texttt{political\_stance\_62}  & train & Farm subsidies distort markets and should be phased out     & Agree & Disagree & \lright \\
6  & \texttt{political\_stance\_68}  & train & Sanctuary city policies undermine federal immigration law   & Agree & Disagree & \lright \\
7  & \texttt{political\_stance\_72}  & train & Homeschooling should face stricter government oversight     & Agree & Disagree & \lleft  \\
8  & \texttt{political\_stance\_76}  & train & Patent terms should be shortened to encourage innovation    & Agree & Disagree & \lnone  \\
9  & \texttt{political\_stance\_104} & train & Peaceful protest is more effective than civil disobedience  & Agree & Disagree & \lnone  \\
10 & \texttt{political\_stance\_145} & train & Billionaires should not exist in a just society             & Agree & Disagree & \lleft  \\
\addlinespace
11 & \texttt{political\_stance\_156} & test  & Financial transaction taxes would reduce market volatility  & Agree & Disagree & \lleft  \\
12 & \texttt{political\_stance\_231} & test  & Foreign aid spending weakens domestic economic priorities   & Agree & Disagree & \lright \\
13 & \texttt{political\_stance\_232} & test  & The US should maintain military bases around the world      & Agree & Disagree & \lright \\
14 & \texttt{political\_stance\_254} & test  & Stricter gun control laws would reduce violent crime rates  & Agree & Disagree & \lleft  \\
15 & \texttt{political\_stance\_266} & test  & Breaking up big tech companies would increase innovation    & Agree & Disagree & \lnone  \\
16 & \texttt{political\_stance\_353} & test  & Campus speech codes violate First Amendment principles      & Agree & Disagree & \lright \\
17 & \texttt{political\_stance\_361} & test  & NATO expansion provokes conflict and should be halted       & Agree & Disagree & \lnone  \\
18 & \texttt{political\_stance\_363} & test  & Defense spending should be reduced to fund social programs  & Agree & Disagree & \lleft  \\
19 & \texttt{political\_stance\_382} & test  & Critical race theory should be taught in K--12 schools      & Agree & Disagree & \lleft  \\
20 & \texttt{political\_stance\_393} & test  & Nuclear power is essential for achieving clean energy goals & Agree & Disagree & \lnone  \\
\bottomrule
\end{tabularx}
\caption{Political stance items, with train/test split and the direction the ``Agree'' pole leans (\lleft{} / \lright{} / neutral).}
\label{tab:political-stance}
\end{table}

\subsection{Label Bias}
\begin{figure}[h!]
    \centering
    \includegraphics[width=0.99\textwidth]{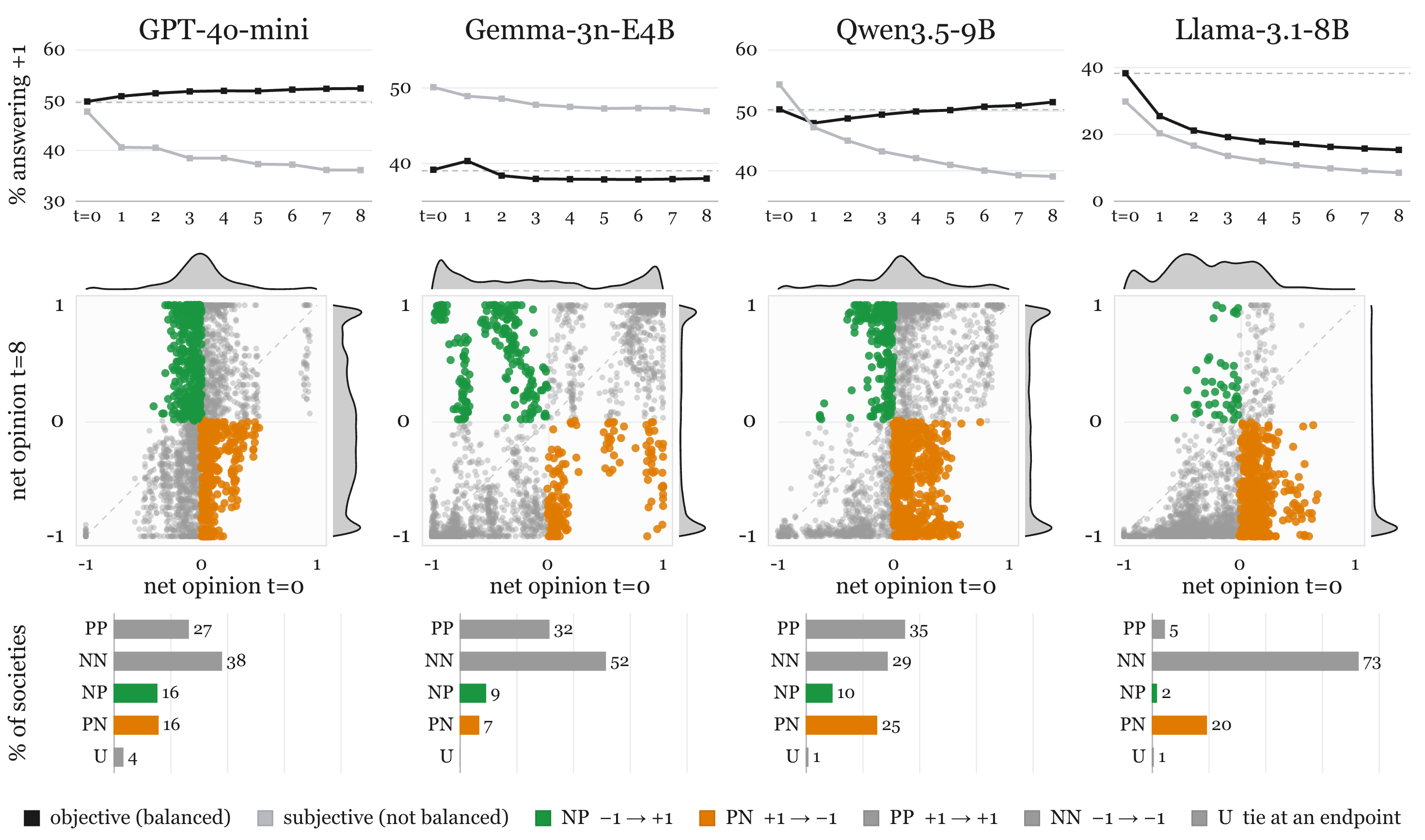}
            \caption{\textit{Label Bias.} How opinions move along the $+1 / -1$ axis over the run ($9,600$ groups from both objective and subjective questions, spanning signed random graphs and the square and triangular lattices, over both the training- and test-set questions). \emph{Row 1:} $y$-axis: percentage of group trajectories, where the net opinion $n(t)$ has positive sign at round $t$. $x$-axis: timesteps. \emph{Row 2:} Each point represents a group trajectory, which is plotted based on its initial ($t=0$, $x$-axis) and final ($t=8$, $y$-axis) net opinion $n(t)$  \emph{Row 3:} Groups switch between $+1$ and $-1$ at $t=0$ and $t=8$. PP (positive$\to$positive), NN (negative$\to$negative), NP (negative$\to$positive), PN (positive$\to$negative), and U (undecided, $50\text{-}50$ tie at $t=0$ or $t=8$). The two transition categories NP and PN are highlighted.}
\label{fig:3_Observations_bias}
\end{figure}

\textit{Label Bias.} A community can also drift toward an answer option irrespective of what that option says (Figure~\ref{fig:3_Observations_bias}). Because the objective questions are label balanced, with the correct answer being $+1$ as often as $-1$, a change in the fraction of agents answering $+1$ over the course of a run cannot be attributed to truth seeking. However, we still observe such a change for Llama-3.1-8B-Instruct, with the fraction answering $+1$ falling from $39\%$ to $16\%$.

\clearpage
\subsection{Thresholds for Characteristic Regimes}\label{apdx:mode_thresholds}

\begin{table}[h!]
\centering
\begin{tabular}{llcc}
\toprule
 & & \multicolumn{2}{c}{Threshold} \\
\cmidrule(lr){3-4}
Regime & Model & Conviction $c^\star$ & Net opinion $n^\star$ \\
\midrule
\multirow{4}{*}{Objective}
              & GPT-4o-mini   & 0.910 & 0.475 \\
              & Gemma-3n-E4B  & 0.940 & 0.994 \\
              & Qwen3.5-9B    & 0.790 & 0.913 \\
              & Llama-3.1-8B    & 0.740 & 0.913 \\
\midrule
\multirow{4}{*}{Subjective}
              & GPT-4o-mini   & 0.960 & 0.375 \\
              & Gemma-3n-E4B  & 0.910 & 0.725 \\
              & Qwen3.5-9B    & 0.820 & 0.600 \\
              & Llama-3.1-8B    & 0.950 & 0.963 \\
\bottomrule
\end{tabular}
\caption{\textit{Selected thresholds for Figure \ref{fig:3_Observations_modes}}. For each model--regime row, the
\emph{conviction threshold} $c^\star$ is defined as the $1/3$ quantile of conviction,
pooled across all episodes, graphs, and rounds; communities with $c(t)<c^\star$ are
classified as \emph{indifferent}. The \emph{net-opinion threshold} $n^\star$ is then
defined as the median of $|n(t)|$ among the remaining ``committed'' communities and
applied symmetrically as $\pm n^\star$ to distinguish \emph{polarization}
($|n(t)|\le n^\star$) from \emph{consensus} ($|n(t)|>n^\star$). Both thresholds are
reported on their native ranges, $c\in[0,1]$ and $|n|\in[0,1]$.}
\label{tab:mode_thresholds}
\end{table}

\subsection{Rare Group Trajectories}\label{apdx:rare_macrostates}

\begin{figure}[h]
    \centering
    \includegraphics[width=0.99\textwidth]{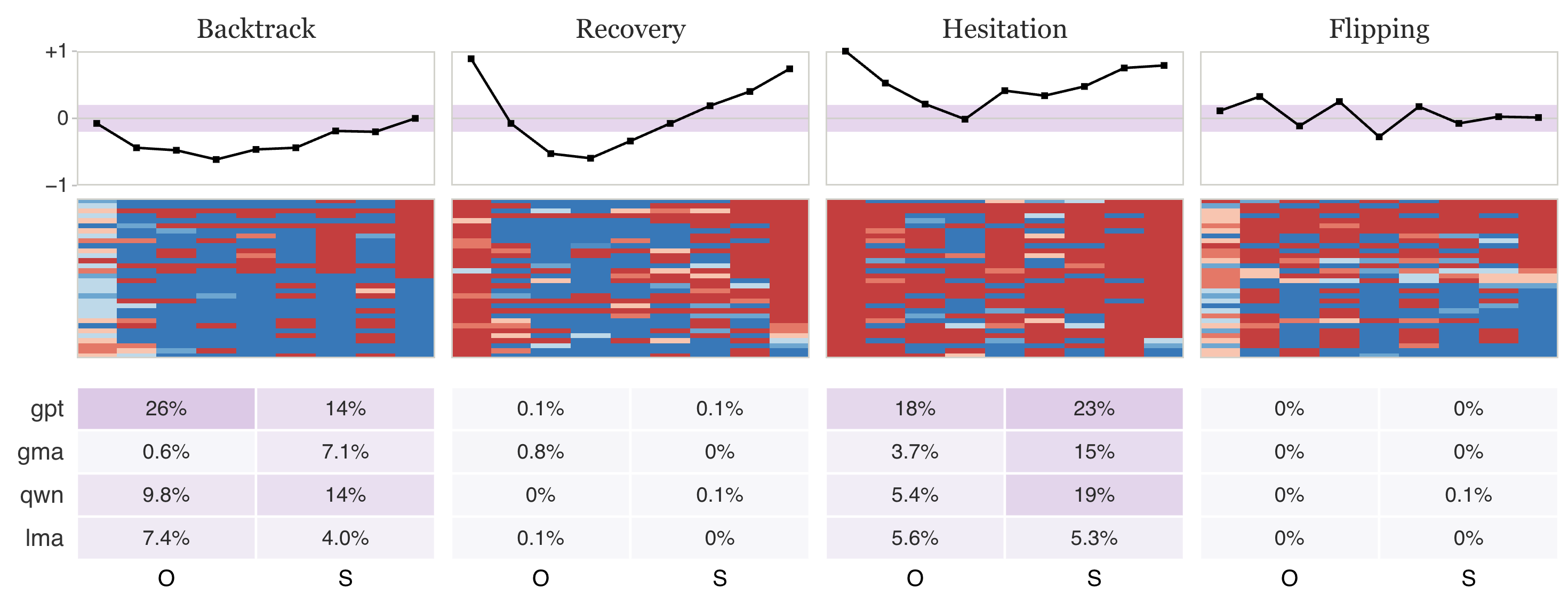}
\caption{\textit{Rare and non-monotone group trajectories.}
Groups whose net opinion \emph{reverses
direction} mid-run, which a first-vs-last-round classification cannot distinguish. Same setup as
Figure ~\ref{fig:macrostates}. A \emph{strong majority} denotes a round with absolute net opinion $|n(t)| \geq 0.5$, which is beyond the
split band $ \le \delta = 0.2$.}
\label{fig:rare_macrostates}
\end{figure} 

\emph{Backtrack}: A split community (absolute net opinion in $\pm 0.2$ split band) first develops a strong majority (absolute net opinion $\geq 0.5$) before relaxing back toward a split. \emph{Recovery}: A strong majority (absolute net opinion $\geq 0.5$) reverses to the opposite side before returning to its original position. \emph{Hesitation}: A strong majority weakens almost to a split, then re-consolidates on the same side. \emph{Flipping}: The majority repeatedly changes sides during the simulation.

\subsection{Effect of Communication Networks on  Trajectories}\label{apdx:communication_effect}

\begin{figure}[h]
    \centering
    \includegraphics[width=0.99\textwidth]{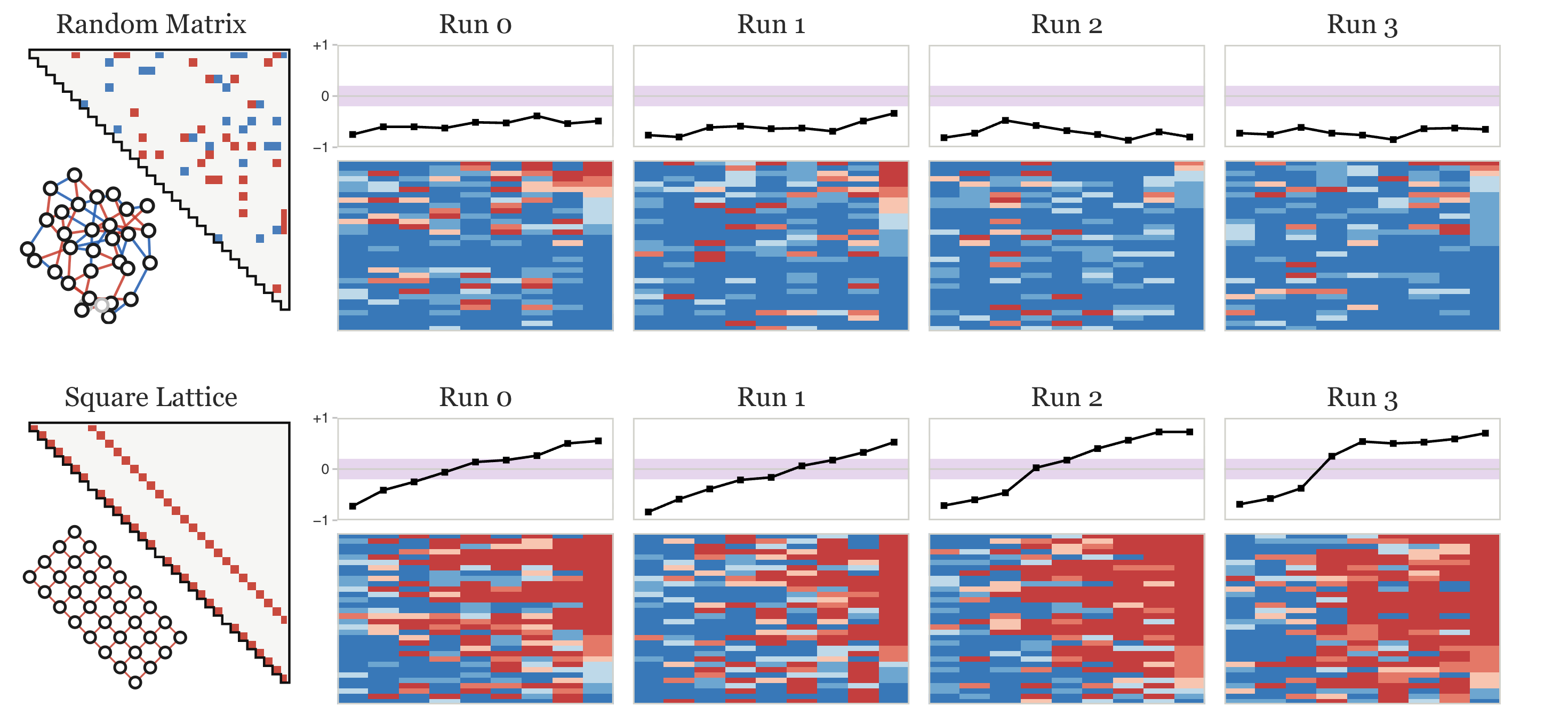}
\caption{\textit{Different communication networks can cause differences in trajectories.} Four episodes (labeled as runs 0-3) on the same question under two different social graphs: a random matrix (\emph{top}) and a square lattice (\emph{bottom}). For each graph, the left panel shows the couplings $J$ as a heatmap in the upper triangle and as a node-link diagram below it, and each of the four runs shows the net opinion $n(t)$ over rounds (\emph{line plot at the top}) together with the group trajectory (\emph{heatmap at the bottom}). Under the random matrix the community stays on its initial side across all four runs, while under the square lattice it crosses the split band and settles on the opposite side in all four runs. This is not a common situation, but social graphs can cause consistent differences between how group trajectories evolve when answering the same question. In other words, the social connections carry information about where will the community end up, even before any messages are exchanged.}
\label{fig:communication_networks}
\end{figure}

\section{Method Details}\label{apdx:method}
\subsection{Local Update Rule}\label{apdx:discrete-updates}

This appendix derives the logistic update rule of Section~\ref{sec:method}.

\textit{Local Field.} Let $s_{i} \in \{+1,-1\}$ be the opinion of agent $i$. Fixing all opinions other than $s_i$, the energy
\[
E(s) \;=\; -\frac{1}{2}\sum_{i=1}^N \sum_{j=1}^N J_{ij}\,s_i s_j \;-\; \sum_{i=1}^N g_i\,s_i
\]
splits into a part that depends on $s_i$ and a part that does not,
\[
E(s) \;=\; -\,s_i f_i \;+\; C,
\qquad
f_i \;=\; \sum_{j\neq i} J_{ij}\,s_j \;+\; g_i ,
\]
where $C$ collects every term independent of $s_i$. The factor $\tfrac12$ disappears because
$J = J^{\top}$ with zero diagonal, so $s_i$ appears twice in the double sum, once in row $i$
and once in column $i$. We call $f_i$ the \emph{local field} on agent $i$: the total pull that
its neighbors and its own predisposition exert on it.

An agent  chooses the next opinion stochastically with temperature $\beta$ based on the energy gap between the two opinion states, $E(s_i{=}{-}1) - E(s_i{=}{+}1) = 2f_i$. We model this with a logistic function of (a) the peer pressure
$\sum_j J_{ij}s_j(t)$ flowing in along the signed graph, scaled by $\tilde\beta$, and (b) an
intrinsic field $\tilde g_i$ encoding the leaning of agent $i$'s persona on this question in
the absence of any peer influence.
\[
P(s_i=+1\mid s_{-i})
\;=\;  \sigma\!\left(2\beta f_i\right) \;=\; \sigma\!\Big(2\beta \big(\textstyle\sum_{j\neq i} J_{ij}\,s_j + g_i\big)\Big)
\;=\; \sigma\!\left(\tilde\beta \sum_{j} J_{ij}\,s_j(t) + \tilde g_i\right),
\]

We omit $\;\tilde{\cdot}\;$ and use $\beta$ and $g_i$ parameters in the main text for simplicity.

\textit{Synchronous Updates.} The observed data record one opinion per agent per round, with
all agents polled at the same round boundary, so we apply the rule synchronously: every agent's
next opinion is drawn from the law above evaluated at the \emph{current} configuration
$s(t)$, and all of them are updated at once to form $s(t+1)$. Appendix~\ref{apdx:continuous-extension} relaxes it by letting agents update at
independent random times, following  \citet{glauber1963}.

\subsection{Continuous Extension}\label{apdx:continuous-extension}

The dynamics of Appendix~\ref{apdx:discrete-updates} assume that all agents update their
opinions synchronously at discrete time steps. We now relax this assumption by introducing a
continuous-time model in which agents revise their opinions independently at random times, as in \citet{glauber1963}.

\subsubsection{Flip Rate}

\textit{Update Rate.} Agent $i$ reconsiders its opinion during a small time window $\mathrm dt$
with probability $\mathrm dt/\tau$, independently of everything that has happened before. We call $1/\tau$ the agent's update rate. Every agent has its own update mechanism, running independently at the rate $1/\tau$.

\textit{Probability of a Flip.} For simplicity, in this section, we move $\beta$ inside $f_i$ and define $f_i \;=\; \beta  \sum_{j\neq i} J_{ij}\,s_j \;+\; g_i$. Note that the probability that $s_i$ is $+1$ is $P(s_i = +1) = \sigma(2f_i)$ and the
probability that $s_i$ is $-1$ is $P(s_i = -1) = 1 - \sigma(2f_i) = \sigma(-2f_i)$. Then, given
the current state is $+1$, the probability of a flip is $\sigma(-2f_i)$, and given the current
state is $-1$, the probability of a flip is $\sigma(2f_i)$. We can write the probability of a
flip more compactly as
\(
P(\text{flip}) = \sigma(-s_i\,2f_i)
\).

\textit{Flip rate $=$ Update Rate $\times$ Probability of a Flip.} Now, we combine the update
rate with the probability of a flip to obtain the flip rate. Using
$\sigma(-2x) = \tfrac{1}{2}\big(1 - \tanh x\big)$ and $\tanh(s_i f_i) = s_i \tanh f_i$, we get
\(
P(\text{flip}) = \sigma(-s_i 2f_i) = \tfrac{1}{2}\big(1 - s_i \tanh f_i\big).
\) The flip rate $w_i$ is the update rate times the flip probability,
\[
w_i(s) = \frac{1}{\tau}\,\sigma(-s_i 2f_i)
= \frac{1}{2\tau}\big(1 - s_i \tanh f_i\big)
= \frac{1}{2\tau}\big(1 - s_i \langle s_i\rangle_c\big).
\]

\subsubsection{Master Equation}

Now, we introduce the master equation, which describes the probability flux between states and
is derived from the conservation of probability mass. At any given time, the system must occupy
one of the possible states in its state space. Consequently, the total probability over all
states must remain equal to one. This implies that probability mass cannot be created or
destroyed. It can only flow from one state to another. Therefore, for each state, the rate of change of its probability is determined by the balance between the probability flux entering the state and the probability flux leaving it. The master equation formalizes this balance by equating the net change in probability to the total inflow minus the total outflow of probability mass.

\textit{Flip operator $\mathcal{F}$.} Define $\mathcal{F}_i s$ as state $s$ with agent $i$ flipped,
\(
\mathcal{F}_i s = [\,s_1,\ \ldots,\ -s_i,\ \ldots,\ s_n\,].
\)

\textit{Master Equation.} The probability of being at state $s$ changes based on the difference
between the inflow into and outflow out of state $s$,
\[
\frac{d}{dt}P(s,t) = \sum_{i=1}^{n}
\underbrace{w_i(\mathcal{F}_i s)\,P(\mathcal{F}_i s, t)}_{\text{inflow into } s}
- \underbrace{w_i(s)\,P(s,t)}_{\text{outflow out of } s}.
\]
This is a probability-flow equation expressing conservation of probability mass in the system,
where $w_i(\mathcal{F}_i s)$ is the probability that $i$ flips at $\mathcal{F}_i s$, and $P(\mathcal{F}_i s, t)$ is the
probability mass allocated to $\mathcal{F}_i s$. The inflow into $s$ is $w_i(\mathcal{F}_i s)\times P(\mathcal{F}_i s, t)$
and the outflow out of $s$ is $w_i(s)\times P(s,t)$, where
\[
w_i(s) = \frac{1}{2\tau}\big(1 - s_i \tanh f_i\big),
\qquad
w_i(\mathcal{F}_i s) = \frac{1}{2\tau}\big(1 - (-s_i)\tanh f_i\big)
= \frac{1}{2\tau}\big(1 + s_i \tanh f_i\big).
\]

\subsubsection{Mean-Field ODE}

The master equation tracks the full distribution $P(s,t)$ over all $2^n$ configurations, which
is intractable to follow directly. Instead, we track only the quantity we care about: each
agent's opinion, 
$m_i(t) = \langle s_i(t)\rangle = \sum_s s_i\,P(s,t)$, which is between
$-1$ (the agent is surely $-1$) and $+1$ (surely $+1$).

We can derive how $m_i$ evolves directly from the master equation. The only events that change $s_i$ are flips of agent $i$ itself. Each flip sends $s_i \mapsto -s_i$, a change of $-2s_i$, and these flips occur at rate $w_i(s)$. Averaging this
instantaneous change over the distribution gives
\[
\frac{d m_i}{dt}
= \big\langle -2\,s_i\,w_i(s)\big\rangle
= -\frac{1}{\tau}\big\langle s_i\big(1 - s_i \tanh f_i\big)\big\rangle
= -\frac{1}{\tau}\Big(\langle s_i\rangle - \big\langle \tanh f_i\big\rangle\Big),
\]
where the last step uses $s_i^2 = 1$. This relation is exact; however, the average
$\langle \tanh f_i\rangle$ is intractable to compute.

\textit{Mean-field approximation.} To make this quantity tractable, we make the mean-field approximation. We assume each agent feels the \emph{average} field of its neighbors rather than their fluctuating instantaneous opinions, and that neighbors are uncorrelated. Consequently, we
can move the average inside the nonlinearity,
\[
\big\langle \tanh f_i\big\rangle \;\approx\; \tanh\big(\langle f_i\rangle\big),
\qquad
\langle f_i\rangle = \beta\sum_{j=1}^n J_{ij}\,\langle s_j\rangle + g_i = \beta\,(Jm)_i + g_i,
\]
replacing each neighbor opinion $s_j \in \{ -1, +1\}$ by the $m_j \in [-1, +1]$. This neglects the correlations between agents, but it reduces the $2^n$-dimensional master equation into a closed system of $N$ ordinary differential equations,
\[
\tau\,\frac{d m_i}{dt} \;=\; -\,m_i \;+\; \tanh\!\Big(\beta\,(Jm)_i + g_i\Big).
\]
Each agent's opinion relaxes, on the timescale $\tau$, toward the response
$\tanh(\beta(Jm)_i + g_i)$ it would settle into given the current average opinions of its
neighbors. The fixed points $m^\star$ are where $\frac{d m}{dt} = 0$ and $m_i^\star = \tanh(\beta(Jm^\star)_i + g_i)$.

\subsubsection{Discretization and the Clock Parameter}

\textit{Discretizing the ODE.} Discretizing the ODE over one observed round of duration
$\mathrm dt$ recovers a form that is similar to the synchronous map of Appendix~\ref{apdx:discrete-updates}. Each agent
updates at least once in the round with probability $\varepsilon = 1 - e^{-\mathrm dt/\tau}$,
\[
m_i(t{+}1) = (1-\varepsilon)\,m_i(t) + \varepsilon\,\tanh\!\Big(\beta\,(Jm(t))_i + g_i\Big),
\]
so $\varepsilon \in (0,1)$ is simply the per-round update rate, with
$\tau/\mathrm dt = -1/\ln(1-\varepsilon)$. Taking $\varepsilon \to 1$ returns the discrete rule,
in which every agent revises its opinion in every round.  The three-coupling split of
Appendix~\ref{apdx:discrete-updates} carries over in the same way, by replacing the argument of the $\tanh$ with the corresponding linear combination of peer sums.

\subsection{Fitting}\label{apdx:fitting}
\paragraph{Features $x$.} All methods share a static field block
$\phi_i = [\,1 \mid p_i \mid q \mid p_i \otimes q\,] \in \mathbb{R}^{16}$,
built from the top-3 PCA scores of the persona embedding ($p_i \in \mathbb{R}^{3}$)
and of the question embedding ($q \in \mathbb{R}^{3}$), their outer product, and a
bias. Persistence and Majority Class involve no fit and no features. The fitted
methods differ only in the columns appended to $\phi_i$.

\textit{Interaction-Free} uses $x = \phi_i \in \mathbb{R}^{16}$ with no
state-dependent columns. Consequently, its predictions are constant across rounds.

\textit{Mean-Field} uses $x = [\,\phi_i \mid \bar{s}(t)\,] \in \mathbb{R}^{17}$,
where $\bar{s}(t) = \frac{1}{n}\sum_j s_j(t)$ is the population-mean opinion.

\textit{Discrete update with one coupling} uses
$x = [\,\phi_i \mid (J \, s(t))_i\,] \in \mathbb{R}^{17}$, where $J$ is the
$n \times n$ interaction matrix and $s(t)$ the vector of size $n$ with current opinions.

\textit{Discrete update with three couplings} uses $x = [\,\phi_i \mid (J^{+} s(t))_i \mid (J^{-} s(t))_i \mid (|J|\, s(t))_i\,]
\in \mathbb{R}^{19}$, the concordant, discordant, and unsigned neighbor interaction terms. 

\textit{Continuous update with one coupling} uses the same design
$x = [\,\phi_i \mid (J \, s(t))_i\,] \in \mathbb{R}^{17}$. Differently, the update is
$m_i = (1-\varepsilon)\, s_i(t) + \varepsilon \tanh(w^{\top}x)$ and the
update rate $\varepsilon$ is an additional parameter (18 parameters in total).

\textit{Continuous update with three couplings} uses
$x = [\,\phi_i \mid (J^{+} s(t))_i \mid (J^{-} s(t))_i \mid (|J|\, s(t))_i\,]
\in \mathbb{R}^{19}$, inside $\tanh(w^{\top}x)$, with $\varepsilon$ again
as an additional parameter.

\paragraph{Target $y$.} The target is the observed next opinion
$y = s_i(t{+}1) \in \{+1,-1\}$, coded as
$\mathbf{1}\{s_i(t{+}1) = +1\} \in \{0,1\}$. We pool all one-step transitions ($t = 0,\dots,7$, all 32 agents) over the train questions, the 8 random training graphs (the lattice and low-rank families are held out entirely), and 4 episodes per configuration.

\paragraph{Loss.} All fitted models maximize a Bernoulli likelihood over the
pooled transitions, with a small $\ell_2$ penalty ($\lambda = 10^{-3}$) on the
persona--question field weights only. The bias, the coupling parameters, and
$\varepsilon$ are unpenalized. The discrete and baseline models use
$P(s_i(t{+}1)=+1) = \sigma(w^{\top}x)$. The continuous model uses
$P(s_i(t{+}1)=+1) = \tfrac{1}{2}(1+m_i)$ with
$m_i = (1-\varepsilon)\, s_i(t) + \varepsilon \tanh(w^{\top}x)$.

\paragraph{Optimization.} All models are fit by full-batch Adam (learning rate $0.05$, $\beta_1 = 0.9$, $\beta_2 = 0.999$, at most 1{,}500 steps with a plateau stop checked every 100). The discrete update models are fitted on standardized feature columns, whereas the continuous update models are fitted on the original feature columns, which we observe perform better than the standardized features. In the continuous model the update rate is reparameterized as $\varepsilon = \sigma(\theta_0)$ to keep it in $(0,1)$. All couplings are initialized at $0.1$ and all other parameters at zero. Because the unsigned drive $(|J|s)_i$ lies in the span of the signed drives, the
three-coupling designs are fit in two stages: $\beta_0$ is estimated first
from the unsigned model, then held fixed as an offset while $\beta^{+}$ and
$\beta^{-}$ are refit.

\subsection{Baselines}\label{apdx:baselines}

We compare the discrete- and continuous-time models against a set of baselines that isolate
the contribution of (i) class imbalance, (ii) temporal persistence, (iii) intrinsic agent
preferences, and (iv) social influence. Writing $d_p$ and $d_q$ for the persona- and
question-embedding dimensions, we report the parameter count of each so that the comparison in
Table~\ref{tab:prediction_models} can be read against model capacity. All baselines with
parameters are fit on the same transitions and with the same objective as the update rules
(Appendix~\ref{apdx:fitting}).

\textit{Majority Class.} Majority class is the simplest possible predictor. It ignores agent identity and interactions and always predicts the majority opinion observed in the training data,
\[
\hat{s}_i(t{+}1)=c,
\qquad
c=\operatorname{sign}\!\Big(\textstyle\sum_{\text{train}} s\Big).
\]
The model contains a single learned parameter, the constant label $c\in\{+1,-1\}$, and measures
the extent to which performance can be explained by class imbalance alone. We report this baseline only in tables that report accuracy (not balanced) because some models exhibit label bias.

\textit{Persistence.} Persistence assumes that opinions do not change and simply copies the
current opinion,
\[
\hat{s}_i(t{+}1)=s_i(t).
\]
Because opinion trajectories are often highly persistent, this baseline can be surprisingly
strong under plain accuracy despite predicting no opinion reversals. For the rollouts, persistence predicts $\hat{s}_i(t)=s_i(0)$.

\textit{Interaction-Free Model.} This baseline retains agent- and question-specific information while removing all social interactions. Predictions depend only on the intrinsic field,
\[
P\big(s_i(t{+}1)=+1\big)
=
\sigma\!\big(g_i\big),
\qquad
g_i={w}_{\text{field}}^{\top}{\phi}_i .
\]
Since the field is independent of the current opinion state, predictions are identical across
rounds and exhibit no dynamics.The model has
$(1+d_p+d_q+d_p d_q)$ learned parameters,\footnote{$d_p = d_q = 3$ and $1+d_p+d_q+d_p d_q = 16$ in all experiments.} the bias, persona and question components,
and their interaction. It isolates how much of an agent's next opinion is predictable
from who it is and what it was asked, before any communication with other agents.

\textit{Mean-Field Model.} This baseline incorporates social influence while discarding network
structure. Instead of interacting through the signed graph, each agent responds only to the
population-average opinion,
\vspace{-4pt}
\[
\bar{s}(t)=\frac{1}{n}\sum_j s_j(t),
\quad \text{yielding} \quad
P\big(s_i(t{+}1)=+1\big)
=
\sigma\!\big(g_i+\beta\,\bar{s}(t)\big).
\]
The model captures a conformity pressure toward the prevailing opinion while ignoring
heterogeneity in social ties. It assumes a $J$ in which every pair of agents is coupled with equal strength. Therefore, this baseline tests whether the signed-network structure provides additional predictive value. Relative
to the Interaction-Free baseline it introduces a single additional coupling parameter $\beta$, for a total of $(1+d_p+d_q+d_p d_q)+1 = 17$ learned parameters.

\subsection{Balanced Accuracy}\label{apdx:metric}

We score predictions with balanced accuracy in Tables \ref{tab:prediction_models} and \ref{tab:generalization}. We sort the observed transitions into four groups, by whether the agent's next opinion flips or stays
($y_i(t{+}1) \neq s_i(t)$ or $y_i(t{+}1) = s_i(t)$) and by the value of that next opinion
($+1$ or $-1$). Balanced accuracy is the unweighted mean of the accuracies on these four groups,
\[
\text{Balanced Accuracy} \;=\; \tfrac{1}{4}\Big(
\mathrm{acc}[\text{flip} \wedge {+}1] +
\mathrm{acc}[\text{flip} \wedge {-}1] +
\mathrm{acc}[\text{stay} \wedge {+}1] +
\mathrm{acc}[\text{stay} \wedge {-}1]
\Big).
\]

We observe that the agent opinions change rarely and the two labels are not equally
common, so the two degenerate strategies both score well under plain accuracy: copying the
current opinion exploits the rarity of flips, and always predicting the more common label
exploits the class imbalance. Balanced accuracy closes both routes at once. A rule that always copies gets
every \textit{stay} group right and every \textit{flip} group wrong, and a rule that always
predicts one label gets both of that label's groups right and both of the other's wrong. In
either case two of the four terms are $1$ and two are $0$. This means a number above $50$ can only be earned by predicting which agents change their minds and in which direction.

\textit{Application to Rollouts.} The rollout columns of Table~\ref{tab:prediction_models}
apply the same four-way decomposition, with the groups defined by the observed transition at
each round and the prediction taken from the rolled-out trajectory
(Appendix~\ref{apdx:rollouts}).

\subsection{Rollouts}\label{apdx:rollouts}

\textit{Deterministic Rollouts.} The rollout numbers reported in parentheses in
Table~\ref{tab:prediction_models} and Table~\ref{tab:generalization} take the modal prediction
at each step,
\[
\hat s_i(t{+}1) = \operatorname{sign}\!\big(h_i(t)\big)
\quad\Longleftrightarrow\quad
\hat s_i(t{+}1) = +1 \ \text{ iff } \ \sigma\big(h_i(t)\big) > \tfrac12 ,
\]
and are scored against the observed $s(t)$ for $t = 1,\dots,T$ with the metric of
Appendix~\ref{apdx:metric}. Thresholding is the right choice when the question is how well the
rule tracks a \emph{particular} trajectory, since sampling would add noise that no predictor
could be expected to match round for round.

\textit{Stochastic Rollouts.} Section~\ref{sec:results_distributions}
asks whether the fitted dynamics produce distribution of group trajectories that look like the
real ones. Hence, we instead sample,
\[
\hat s_i(t{+}1) = +1 \quad\text{with probability}\quad \sigma\big(h_i(t)\big),
\]
independently across agents and rounds, and compare the resulting distribution of individual and
group archetypes against the real one (Figure~\ref{fig:macrostate_distribution}). Both
variants use the identical fitted rule and differ only in this last step. The temperature sweep experiments also use stochastic rollouts with $1/\mathcal{T}$ scaling $h_i$: $\sigma\big(h_i(t) /\mathcal{T}\big)$.

\newpage
 
\section{Additional Results}\label{apdx:results} 
\subsection{Raw Accuracy and the Utility of the Continuous Model}
\label{apdx:res_utility_of_continuous}

Even when predicting agent communities that update synchronously in lockstep, the continuous rule outperforms the discrete rule when evaluated by accuracy. This difference disappears under flip-balanced accuracy because, for one-step predictions, the continuous rule adds the discrete rule with a fitted flip threshold. Because flips are rare in the training data, the fitted $\varepsilon$ trades sensitivity to flips for precision on the far more common non-flip transitions. Raw accuracy rewards this, whereas balanced accuracy reweights flips and non-flips to $50/50$ and neutralizes it, which is why we report raw accuracy in these tables.
\begin{table}[h!]
\centering
\setlength{\tabcolsep}{3pt}
\resizebox{\textwidth}{!}{%
\begin{tabular}{l cc cc cc cc}
\toprule
 & \multicolumn{2}{c}{GPT-4o-mini} & \multicolumn{2}{c}{Gemma-3n-E4B} & \multicolumn{2}{c}{Qwen3.5-9B} & \multicolumn{2}{c}{Llama-3.1-8B} \\
\cmidrule(lr){2-3}\cmidrule(lr){4-5}\cmidrule(lr){6-7}\cmidrule(lr){8-9}
Method & In-d. & Out-d. & In-d. & Out-d. & In-d. & Out-d. & In-d. & Out-d. \\
\midrule
Majority Class & 64.0 (64.0) & 61.9 (61.9) & 54.0 (54.0) & 55.9 (55.9) & 50.5 (50.5) & 49.2 (49.2) & 87.2 (87.2) & 87.4 (87.4) \\
Persistence & 76.9 (73.2) & 73.7 (71.6) & 81.3 (70.9) & 81.1 (70.6) & 75.3 (75.2) & 74.3 (\textbf{74.8}) & 85.1 (75.3) & 85.6 (74.4) \\
Interaction-Free & 54.0 (54.0) & 51.0 (51.0) & 64.3 (64.3) & 62.5 (62.5) & 48.4 (48.4) & 46.3 (46.3) & 49.6 (49.6) & 48.9 (48.9) \\
Mean-Field & 67.6 (59.1) & 66.6 (60.4) & 78.3 (73.5) & 77.5 (\textbf{74.8}) & 73.6 (71.7) & 73.9 (73.1) & 86.3 (74.9) & 86.2 (73.9) \\
\midrule
Discrete Update & 76.5 (68.7) & 74.4 (67.3) & 64.7 (64.1) & 62.5 (62.0) & 63.3 (56.6) & 58.0 (53.6) & 55.2 (53.7) & 53.8 (52.9) \\
\quad + Three Couplings & 85.9 (76.9) & 85.4 (76.7) & 80.9 (73.9) & 79.7 (70.4) & 83.3 (75.1) & 80.3 (71.9) & 90.6 (80.8) & 90.0 (80.8) \\
\midrule
Continuous Update & 82.7 (73.0) & 80.4 (70.3) & 81.2 (67.4) & 80.3 (65.5) & 74.5 (58.3) & 70.7 (55.0) & 75.0 (61.2) & 73.8 (60.2) \\
\quad + Three Couplings & \textbf{89.3} (\textbf{80.0}) & \textbf{87.6} (\textbf{80.3}) & \textbf{84.5} (\textbf{76.6}) & \textbf{83.6} (72.4) & \textbf{85.4} (\textbf{75.6}) & \textbf{82.4} (73.4) & \textbf{94.2} (\textbf{88.2}) & \textbf{93.7} (\textbf{88.0}) \\
\bottomrule
\end{tabular}%
}\caption{\textit{Predictions on Subjective Questions.} Same setup as Table \ref{tab:prediction_models} but uses the raw accuracy as a metric instead of balanced accuracy.\label{tab:prediction_models_sub}}

\end{table}

\begin{table}[h!]
\centering
\setlength{\tabcolsep}{3pt}
\resizebox{\textwidth}{!}{%
\begin{tabular}{l cc cc cc cc}
\toprule
 & \multicolumn{2}{c}{GPT-4o-mini} & \multicolumn{2}{c}{Gemma-3n-E4B} & \multicolumn{2}{c}{Qwen3.5-9B} & \multicolumn{2}{c}{Llama-3.1-8B} \\
\cmidrule(lr){2-3}\cmidrule(lr){4-5}\cmidrule(lr){6-7}\cmidrule(lr){8-9}
Method & In-d. & Out-d. & In-d. & Out-d. & In-d. & Out-d. & In-d. & Out-d. \\
\midrule
Majority Class & 47.2 (47.2) & 46.5 (46.5) & 58.6 (58.6) & 58.5 (58.5) & 49.1 (49.1) & 49.4 (49.4) & 79.4 (79.4) & 77.9 (77.9) \\
Persistence & 63.2 (58.6) & 62.4 (57.7) & 86.6 (78.6) & 87.1 (79.3) & 81.8 (63.9) & 82.4 (63.5) & 75.7 (66.2) & 75.8 (65.6) \\
Interaction-Free & 48.9 (48.9) & 47.9 (47.9) & 61.6 (61.6) & 62.0 (62.0) & 43.5 (43.5) & 44.7 (44.7) & 79.4 (79.4) & 77.9 (77.9) \\
Mean-Field & 69.5 (56.4) & 69.3 (56.6) & 90.2 (\textbf{82.9}) & 90.4 (\textbf{84.2}) & 87.3 (\textbf{69.6}) & 87.6 (\textbf{68.6}) & 83.2 (75.0) & 82.2 (72.4) \\
\midrule
Discrete Update & 68.8 (60.7) & 66.2 (59.3) & 61.1 (61.2) & 61.6 (61.9) & 47.9 (46.3) & 47.6 (47.1) & 79.3 (79.0) & 77.6 (77.3) \\
\quad + Three Couplings & 86.7 (67.4) & 85.5 (64.9) & 91.6 (81.8) & 91.5 (81.5) & 91.3 (66.7) & 90.9 (65.4) & 89.3 (78.8) & \textbf{88.4} (77.5) \\
\midrule
Continuous Update & 70.9 (57.9) & 68.3 (56.2) & 86.7 (69.2) & 87.1 (69.8) & 73.8 (47.2) & 73.3 (48.2) & 79.7 (79.4) & 78.1 (77.7) \\
\quad + Three Couplings & \textbf{87.2} (\textbf{70.0}) & \textbf{85.6} (\textbf{67.8}) & \textbf{92.4} (82.0) & \textbf{92.6} (81.9) & \textbf{91.5} (67.1) & \textbf{91.1} (65.4) & \textbf{89.4} (\textbf{79.8}) & \textbf{88.4} (\textbf{78.3}) \\
\bottomrule
\end{tabular}%
}
\caption{\textit{Predictions on Objective Questions.} Same setup as Table \ref{tab:prediction_models} but uses the raw accuracy as a metric instead of balanced accuracy.\label{tab:prediction_models_obj}}
\end{table}

\newpage 
\subsection{Asynchronous-update experiment}\label{apdx:async_update}

In this appendix, we report the results of our experiments where  individual agents update with independent update rates asynchronously. 

\textit{Generating the update schedule.} We use the same questions, the same 32 agents, and the same random interaction graphs as the main experiment. For each question–$J$ pair, we generate one trajectory and run it for $T=8$ timesteps. Here, we divide each step into $1000$ equal cells, so one cell is $1/1000 $ of a step. In each cell, each agent updates with a probability of 0.5/1000, independently of all other agents and cells. This means each agent updates 0.5 times per step on average (4 times over the whole run), and the times between its updates are random. Reading the sampled cells in order gives, for every step, the list of who updates and in what order. In the rare case where two agents land in the same time cell, we put them in a random order, so no two agents ever update at the same moment. 

\textit{Running the dynamics.} At time $0$, every agent answers the question with an empty inbox, exactly as in the synchronous experiment. After that, the run follows the saved schedule. When an agent's turn comes, it does what an agent does in one synchronous round: it reads its current inbox (the most recent message from each neighbor), writes a message stating its current answer, sends it to its neighbors, and then samples a new answer from the model (majority vote over $K = 5$ samples, as before). Because updates happen one at a time, an agent that updates late in a step sees the new messages of agents that updated earlier in the same step. This is the key difference from the synchronous setting, where all agents read the previous round's messages.

\begin{table}[h]
\centering
\small
\begin{tabular}{l cccc cc cc}
\toprule
& \multicolumn{4}{c}{Baselines}
& \multicolumn{2}{c}{Discrete Update}
& \multicolumn{2}{c}{Continuous Update} \\
\cmidrule(lr){2-5} \cmidrule(lr){6-7} \cmidrule(lr){8-9}
& M & P & IF & MF
& Base & + Three Couplings
& Base & + Three Couplings \\
\midrule
Subjective In-d.  & 61.1 & 78.3 & 56.3 & 63.4 & 60.6 & 66.0 & 79.7 & \textbf{80.4} \\
Subjective OOD    & 58.9 & 76.9 & 52.2 & 59.8 & 57.7 & 62.8 & 77.3 & \textbf{78.5} \\
Objective In-d.   & 46.7 & 64.3 & 48.9 & 52.8 & 51.4 & 54.9 & 63.8 & \textbf{65.3} \\
Objective OOD     & 48.2 & 65.0 & 48.0 & 55.1 & 53.0 & 54.7 & 64.7 & \textbf{65.9} \\
\bottomrule
\end{tabular}
\caption{\textit{Prediction under asynchronous dynamics.} The baselines are Majority Class (M), Persistence (P), Interaction-Free (IF), and Mean-Field (MF). The table reports the rollout accuracies. The model is fitted and evaluated on the asynchronous setup described above. GPT-4o-mini is used as the language model. \label{tab:async}}
\end{table}

We observe that the continuous model outperforms the discrete model with a larger margin under this setup.

\newpage 
\subsection{Temperature Sweep Experiment}\label{apdx:per_society_transitions}
\label{apdx:temperature_sweep}

\begin{figure}[h]
    \centering
    \includegraphics[width=0.99\textwidth]{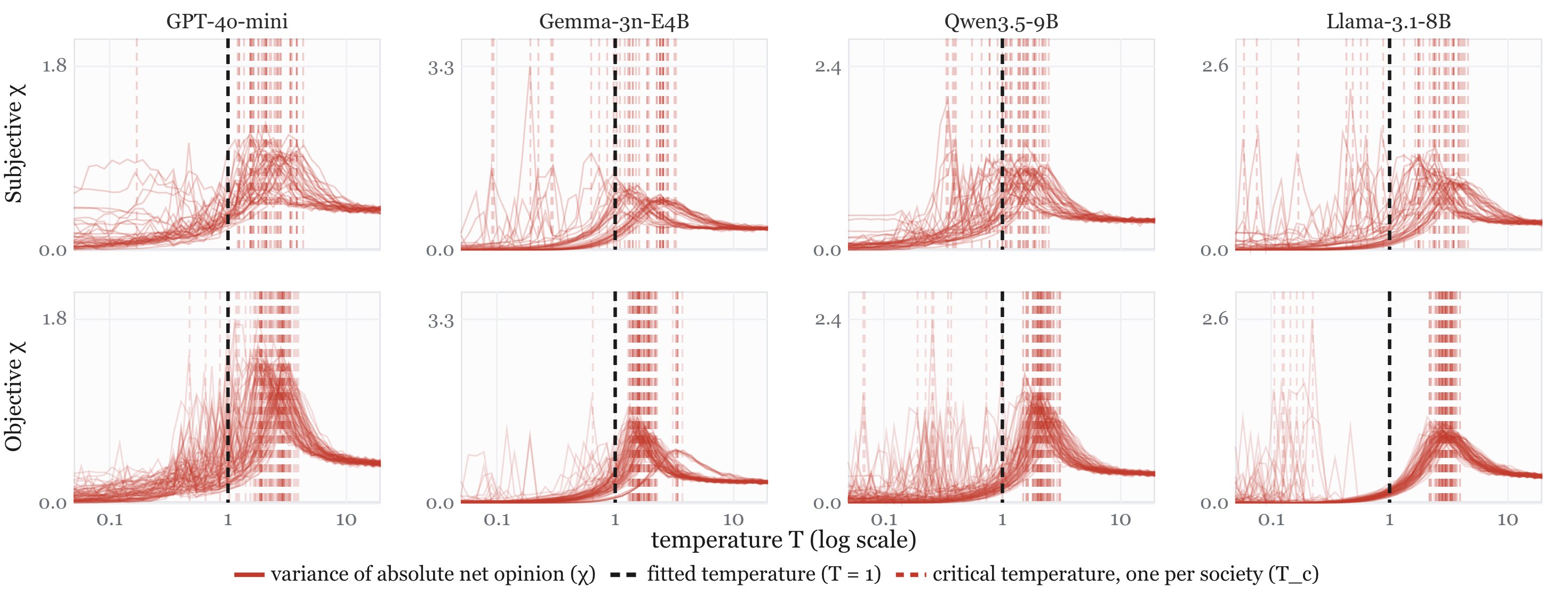}
\caption{\textit{Figure~\ref{fig:agent_model_tc_mean_test_J03_fitted_nomf_tcmean} with each question-$J$ pair presented individually.} Figure~\ref{fig:agent_model_tc_mean_test_J03_fitted_nomf_tcmean} in
Section~\ref{sec:results_temperature} reports the across-community mean of these curves and the
across-community mean of their $\chi$-peak $\mathcal{T}_c$'s. This figure shows the same sweep without any
averaging with one curve and one $\mathcal{T}_c$ line per group. \label{fig:agent_model_tc_mean_test_J03_fitted_nomf_tcmean_qea}}
\end{figure}

This appendix explains how the critical temperatures in Section~\ref{sec:results_temperature} are computed. To avoid confusion, we begin by clarifying that this experiment does not modify the language model’s temperature. Instead, it modifies the temperature of the fitted model, as described below.

\textit{Setup.} The experiment covers every question in the test set and four in-distribution random graphs. Four graphs $\times$ the test question bank gives $80$
objective and $40$ subjective groups per model, each with four independent episodes. 

\textit{Model.} The sweep uses the parameters of the three-coupling discrete update rule.

\textit{Introducing Temperature.} Our fitted rule is
$P\big(s_i(t{+}1)=+1\big) = \sigma\big(h_i\big)$. Reintroducing the temperature divides that
logit by $\mathcal{T}$,
\[
P\big(s_i = +1\big) \;=\; \sigma\!\left(\frac{h_i}{\mathcal{T}}\right),
\]
so the $\mathcal{T}=1$ point of the sweep is the fitted model itself.

\textit{Initialization.} We call a $J$ paired with a question a \textit{group}. For each group, we have taken $4$ independent trajectories in our experiments. The $4$ trajectories are considered as the different rollouts of the same system in the temperature sweep as well. We initialize a group at each of the four starting points at $t=0$ from the $4$ independently sampled trajectories.  

\textit{Procedure.} We sweep $41$ log-spaced temperatures from $0.05$ to $20$.\footnote{$0.050, 0.058, 0.067, 0.078, 0.091, 0.106, 0.123, 0.143, 0.166, 0.193, 0.224, 0.260, 0.302, 0.350, 0.407, 0.473, 0.549, 0.638, 0.741, 0.861, 1.000,\\ 1.162, 1.349, 1.567, 1.821, 2.115, 2.456, 2.853, 3.314, 3.850, 4.472, 5.195, 6.034, 7.009, 8.142, 9.457, 10.986, 12.761, 14.823, 17.218, 20.000
$} At each
temperature, we run the model for $500$ synchronous steps. We discard the first $200$ (to account for the time it takes for the community to reach equilibrium) and retain the last $300$. During this run, each agent's opinion in the next step $s_i = \pm1$ is sampled with probability $\sigma(h_i/\mathcal{T})$.

\textit{Observables.} From the $300$ steps of the group trajectories, we compute $\chi \;=\; N\,\mathrm{Var}_t\big(|n(t)|\big)$. The four independent trajectories per group are used as four independent initial conditions that are then averaged into one $\chi$ curve per group.

\textit{Locating $\mathcal{T}_c$.} A group's $\mathcal{T}_c$ is the location of its $\chi$ peak. Since the grid is
coarse relative to the width of the peak, we fit a parabola through the maximizing
grid point and its two neighbors in $\log \mathcal{T}$ and take the vertex, which resolves $\mathcal{T}_c$ below the
spacing of the $41$-point grid.

\subsection{Predictability of Group Archetypes}\label{apdx:replica_predictability}

\begin{figure}[h]
    \centering
    \includegraphics[width=0.99\textwidth]{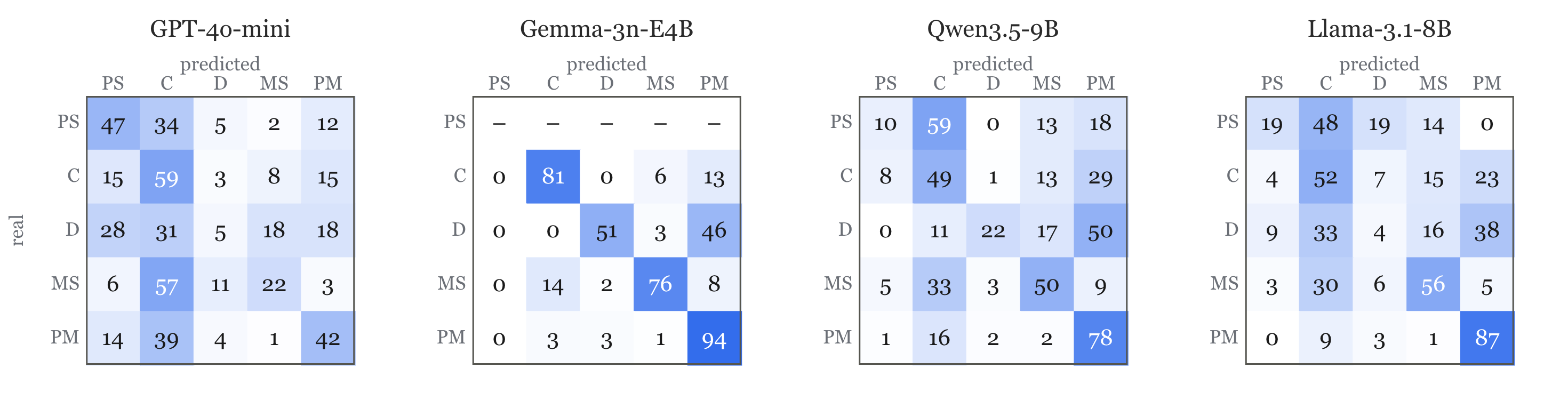}
    \includegraphics[width=0.99\textwidth]{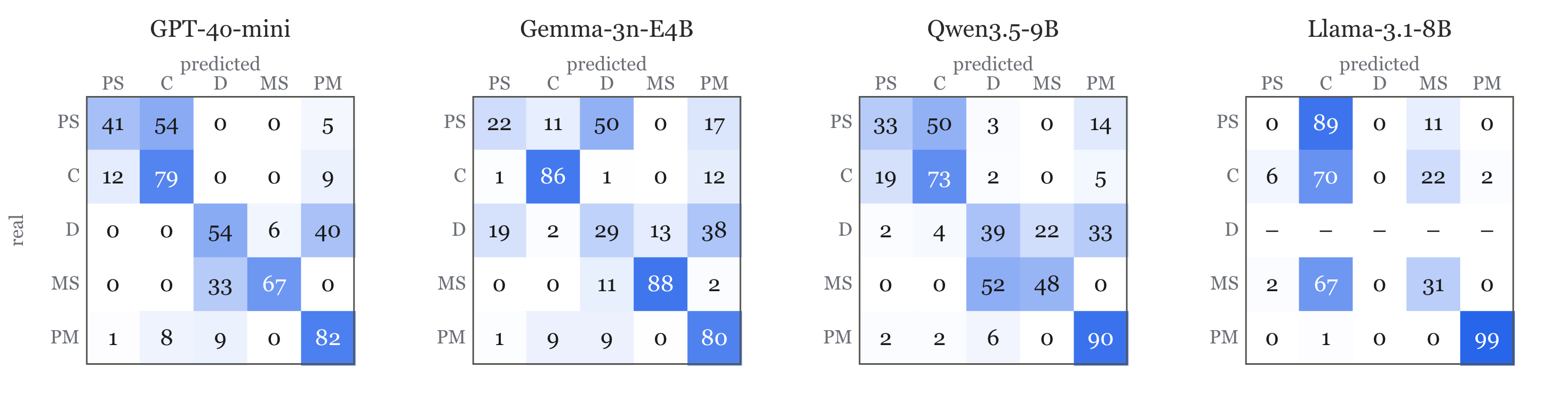}
\caption{\textit{Predictability of the Group Archetypes (row 1: objective; row 2: subjective).} For every (question $\times$ graph) pair, we assess agreement among the $4$ independently sampled episodes. For this analysis, we use test questions and $4$ random graphs as our $J$s. For each question--$J$ pair, we use the group archetype of one of the $4$ trajectories as the predicted group archetype. We then use the remaining $3$ group trajectories as the test set to construct a confusion matrix for predicting the correct group archetype. We repeat this process, using each of the $4$ trajectories in turn as the prediction and the remaining $3$ as the test set, in a cross-validation manner, and report the \textit{``cross-validated''} confusion matrix. Because every trajectory takes its turn as the predictor, the cross-validated matrix is symmetric in raw counts. Row normalization makes the confusion matrix asymmetric. A row of the confusion matrix shows how often, for example, a \textit{persistent split} trajectory is predicted as each of the other group archetypes. If the repeated samples always matched, we would observe a diagonal confusion matrix with $100$ on the diagonal. \label{fig:predictability_macro_subjective_test_J03_cv_row}}\end{figure}

\newpage
\subsection{Flow of Opinion} 
\label{apdx:f1}

\begin{figure}[h]
    \centering
\includegraphics[width=0.99\textwidth]{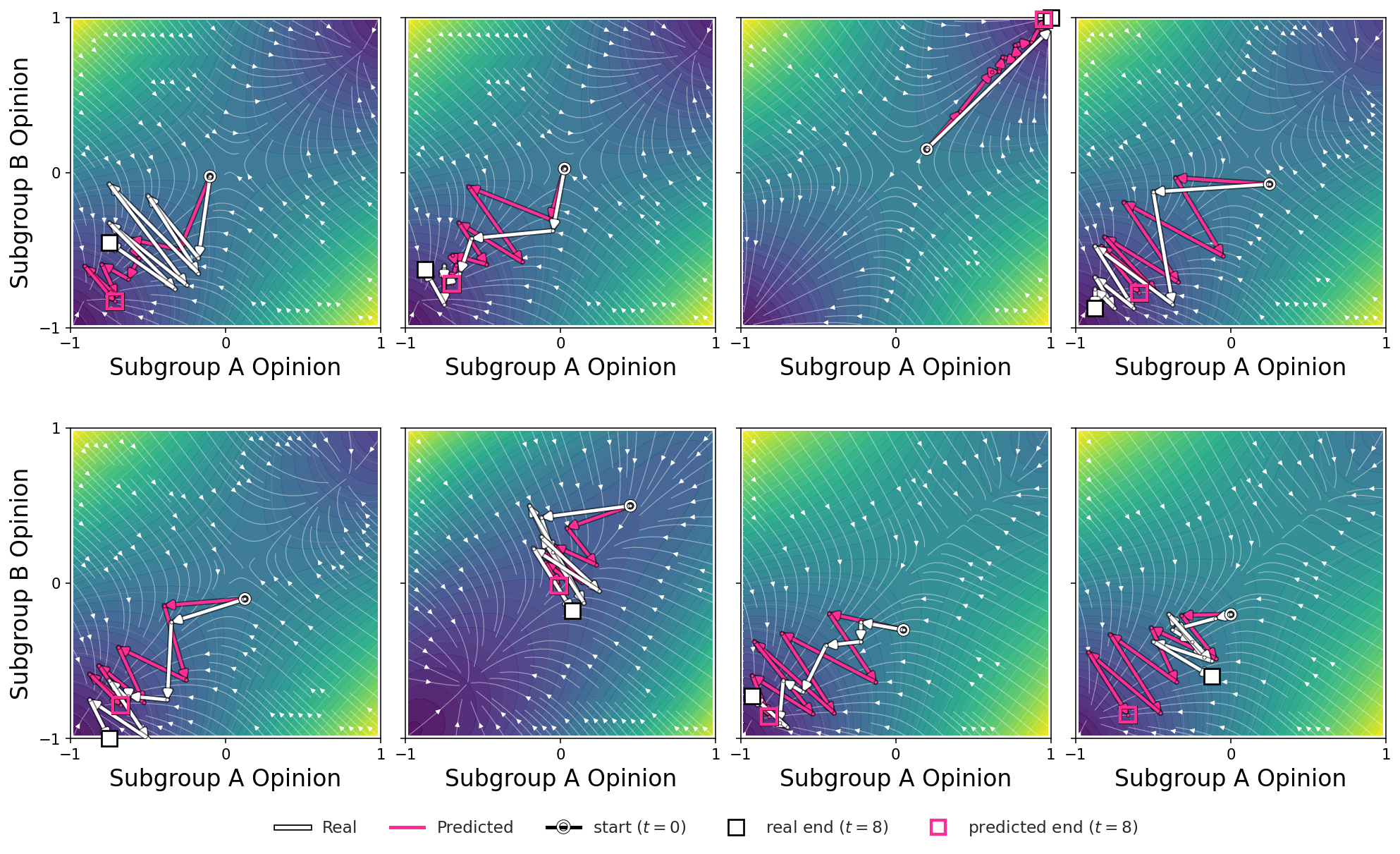}
\caption{\textit{Flow of opinion predicted by the continuous three coupling model.} Each panel is one
group of 32 language-model agents interacting about a question over eight rounds. The axes are the average opinions of the two subgroups. The background shows our fitted model's predictions, with energy represented by colors and the predicted flow (at the continuous time limit) represented by streamlines in the background. White indicates the observed transitions, while pink shows the model's predicted rollout from the initial opinions at $t=0$, that updates the agents synchronously at discrete time intervals $t=1,\dots, T$.  
}
\end{figure}
\begin{figure}[h!]
    \centering
\includegraphics[width=0.99\textwidth]{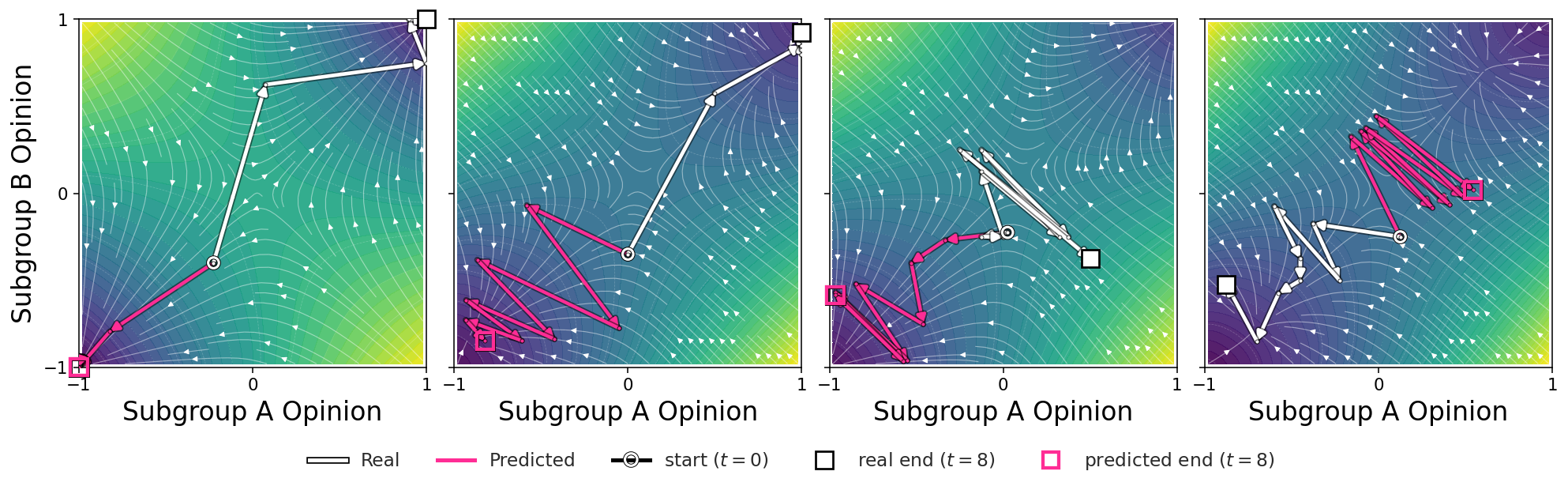}
\caption{\textit{Four example group trajectories where predictions do not match the observations.}
}
\end{figure}
\textit{Splitting into groups.} We split the agents into two subgroups based on their opinion at $t=0$. The $N=32$ agents are ranked by their round-0 opinion $\bar{o}_i(0)$; the top half forms group $A$ and the bottom half group $B$, so the two
subgroups always have equal size ($16$ and $16$) and $A$ is by construction the side that starts higher, $n_A(0)\ge n_B(0)$, where $n_A(0)$ denotes the net opinion of the agents in subgroup A. Ties are broken by the following round, then the one after.

At each point on a $220\times220$ grid, we record the net opinions of agents in $A$, $n_A(t)$, and agents in $B$, $n_B(t)$. We generate rollouts and plot the subgroup net opinions at each timestep. The white path shows the real community’s per-round net opinions in the same two-dimensional space, while the pink path shows a rollout from the fitted continuous three-coupling update rule, initialized from the same state. We show selected rollouts on the held-out test set of questions.

\subsection{Experiments with Frontier Models and Larger Mixed Groups}\label{apdx:frontier_mixed}

\paragraph{Setup.}
To test whether the fitted update rules transfer to a larger, mixed-model
group, we instantiate a single group of $n=64$ agents from the same bank
of 32 personas used in the main experiments. In this experiment, each persona appears twice, once
served by GPT-5.6-sol \citep{openai2026gpt56} (agents $1\dots32$) and once by DeepSeek-V4-Flash \citep{deepseekai2026deepseekv4} (agents
$33\dots64$), so the two model families share identical persona compositions and
differ only in the underlying language model. All episodes run on one fixed signed random
graph (224 non-zero entries, average degree $3.5$ as in the $n=32$
runs), with the same interaction protocol as Table \ref{tab:prediction_models}, $T=8$ timesteps, a single
spin sample per agent per step ($k=1$), and one episode per question over all
40 objective and 20 subjective questions. We fit on the train questions and report accuracy on the
held-out test questions (20/20 objective, 10/10 subjective) on the same graph.

\paragraph{Results.} Table~\ref{tab:frontier_balanced} reports balanced accuracy used in the main text (Section~\ref{sec:results_prediction}) over all agents (All) and separately for the agents backed by each model. As in Table~\ref{tab:prediction_models}, one-step accuracy is reported first and rollout accuracy in parentheses. The discrete update with three couplings remains the best method under balanced accuracy for both question types and for both model families, reaching $81.9$ one-step balanced accuracy on subjective and $80.0$ on objective questions over all agents. The model captures the collective dynamics of a heterogeneous community.

\begin{table}[h]
\centering
\small
\begin{tabular}{l ccc}
\toprule
Method & All & GPT-5.6-sol & DeepSeek-V4-Flash \\
\midrule
\multicolumn{4}{l}{\textit{Subjective Questions}} \\
\midrule
Persistence & 50.0 (63.3) & 50.0 (63.2) & 50.0 (63.5) \\
Interaction-Free & 51.7 (51.7) & 52.2 (52.2) & 51.1 (51.1) \\
Mean-Field & 57.3 (56.3) & 57.0 (56.7) & 57.6 (56.0) \\
Discrete Update & 63.3 (59.0) & 65.5 (60.4) & 61.2 (57.4) \\
+ 3 Couplings & \textbf{81.9 (70.4)} & \textbf{83.9 (71.1)} & \textbf{79.8 (69.7)} \\
\midrule
\multicolumn{4}{l}{\textit{Objective Questions}} \\
\midrule
Persistence & 50.0 (57.7) & 50.0 (62.8) & 50.0 (53.2) \\
Interaction-Free & 48.4 (48.4) & 49.2 (49.2) & 48.2 (48.2) \\
Mean-Field & 75.0 (70.9) & 72.2 (68.7) & 76.2 (71.5) \\
Discrete Update & 48.4 (47.9) & 49.0 (48.8) & 48.2 (47.5) \\
+ 3 Couplings & \textbf{80.0 (67.4)} & \textbf{74.5 (64.6)} & \textbf{83.5 (69.2)} \\
\bottomrule
\end{tabular}
\caption{\textit{Prediction with GPT-5.6-sol and DeepSeek-V4-Flash agents.} Balanced accuracy (one-step balanced accuracy and rollout balanced accuracy in parentheses) of predictions on the held-out test questions of the mixed-family run ($n = 64$, one random graph used for both training and test).\label{tab:frontier_balanced}}
\end{table}

\stopcontents[apx]
\newpage

\end{document}